\documentclass{article}
\usepackage{iclr2027_conference,times}

\usepackage{amsmath,amsfonts,bm}

\def\eqref#1{equation~\ref{#1}}

\def\1{\bm{1}}

\DeclareMathAlphabet{\mathsfit}{\encodingdefault}{\sfdefault}{m}{sl}
\SetMathAlphabet{\mathsfit}{bold}{\encodingdefault}{\sfdefault}{bx}{n}

\usepackage{amsmath}
\usepackage{amssymb}
\usepackage{mathtools}
\usepackage{graphicx}
\usepackage{booktabs}
\usepackage{longtable}
\usepackage{tabularx}
\usepackage{array}
\usepackage[nopatch=eqnum]{microtype}
\usepackage{enumitem}
\usepackage{placeins}
\usepackage{needspace}
\usepackage{caption}
\usepackage{xcolor}
\usepackage[pagebackref=true]{hyperref}
\usepackage{url}
\hypersetup{hidelinks, pdftitle={Velocity Scaling in Flow Matching}, pdfauthor={Youssef Saied and François Fleuret}}

\newcolumntype{L}[1]{>{\raggedright\arraybackslash}p{#1}}

\setlist{itemsep=0.1em,topsep=0.25em,parsep=0pt,partopsep=0pt}

\title{Velocity Scaling in Flow Matching}

\author{%
Youssef Saied \\
University of Geneva \\
Geneva, Switzerland \\
\texttt{youssef.saied@unige.ch}
\And
Fran\c{c}ois Fleuret \\
University of Geneva and Meta FAIR \\
Geneva, Switzerland
}

\iclrfinalcopy

\begin{document}
\raggedbottom{}

\maketitle
\fancyhead{}
\renewcommand{\headrulewidth}{0pt}


\begin{abstract}
    Scaling a learned flow-matching velocity field $v_\theta$ by a gain $\gamma(t)$ was recently shown to greatly improve generation quality.
    Prior work argued that velocity fields trained with mean-squared error (MSE) systematically underestimate velocity magnitude and that scaling corrects this error.
    We show that MSE training does not create a velocity-magnitude deficit.
    We find instead that velocity scaling reduces \textit{population time lag}: sampled states at model time $t$ resemble training states from an earlier time.
    Velocity scaling and moving model time back are two ways to address this population time lag.
    Across architectures and model sizes, measuring population time lag and using it to select a gain greatly improves generation quality, reducing FID from 28.0 to 12.2\footnote{\label{fn:fid-interpolation}Estimated by linear interpolation between FID measurements at neighboring gains.} on ImageNet-256 at NFE 25 without guidance.
\end{abstract}

\section{Introduction}

Flow matching trains a time-dependent velocity field that transports a source distribution to the data distribution~\citep{lipman2023flowmatching}.
Samples are generated by numerically integrating this velocity field, so generation depends not only on the learned field but also on how it is integrated.
Recently, \citet{li2026velocitydeficit} showed that multiplying a learned velocity field \(v_\theta\) by a gain \(\gamma(t)\) during sampling improves generation quality: for unguided SiT-XL/2~\citep{ma2024sit} at NFE 50,\footnote{NFE is the number of function evaluations.} Fr\'echet Inception Distance (FID\@; lower is better)~\citep{heusel2017fid} fell from 13.7 to 7.6.

\citet{li2026velocitydeficit} attribute this improvement to a systematic velocity deficit caused by MSE training.
We show that MSE itself causes no such deficit: the velocity field that minimizes MSE transports the source distribution to the data distribution exactly.
A separate possibility is that finite training leaves the learned field systematically underscaled.
For pretrained SiT, however, the scalar multiplier that best matches predicted velocities to target velocities stays near one at every time and averages 0.999 (Section~\ref{sec:no-deficit}).
That leaves the central question: why does velocity scaling improve sampled images?

We find that sampled states exhibit \emph{population time lag}: states generated at model time $t$ resemble genuine training states from an earlier time.
This lag can be measured directly without a generation-quality metric and increases as NFE decreases (Section~\ref{sec:catchup}).
Related methods address this state--time mismatch by training the model on states closer to those encountered during sampling~\citep{li2025mixflow}, or by changing the time supplied to the model~\citep{li2024timeshift}.

\begin{figure}[t]
    \centering
    \includegraphics[width=\textwidth]{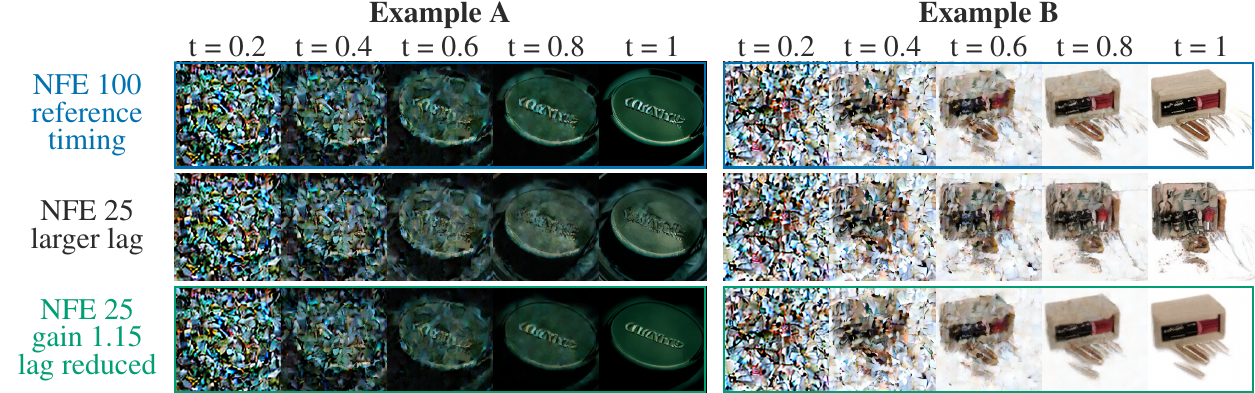}
    \caption{
        Trajectories sampled with an unguided stochastic Euler sampler.
        Within each example, all rows share the same class label, initial latent, and Brownian path.
        The coarse unscaled path develops the object later in the sequence than the finer path; constant gain 1.15 shifts the coarse NFE-25 trajectory toward the finer timing.
        Among the tested gains, a sweep with 50,000 images per gain found the lowest FID at gain 1.15 for NFE 25.
    }%
    \label{fig:visual-clock}
\end{figure}

Moving model time backward adapts the time input to a lagging state; velocity scaling advances the state toward the time supplied to the next model call.
We show that, to first order, velocity scaling advances a lagging state, and empirically find that it eliminates most of the measured lag (Section~\ref{sec:catchup}).
Velocity scaling makes objects develop earlier along coarse sampling trajectories, closer to their timing under finer sampling (Figure~\ref{fig:visual-clock}).
Using the measured lag to choose the gain also sharply improves image quality: at NFE 25, the resulting \emph{lag-based gain} gives an estimated FID of 12.2\footnotemark[\getrefnumber{fn:fid-interpolation}], down from 28.0 (Section~\ref{sec:lag-based-gain}).

A gain sweep lowers FID further to 8.7 at gain 1.15, the \emph{lowest-FID gain} among those tested.
To test whether the two gains follow a simple relation across NFEs, we fit a multiplicative factor $\kappa$ per model using their values at NFE 25.
Reusing this $\kappa$ at NFE 50, 100, and 250 shows that one fixed factor per model describes the additional gain beyond lag correction reasonably well (Section~\ref{sec:gain-settings}).

\Needspace{7\baselineskip}
Appendix~\ref{app:questions-findings} summarizes the questions, main findings, and where to find the supporting evidence.
Our contributions are:
\begin{itemize}[leftmargin=1.5em]
    \item \textbf{Analysis:} Velocity scaling does not fix an underscaled network: MSE training preserves population transport, and the gain that minimizes MSE averages 0.999 for pretrained SiT (Section~\ref{sec:no-deficit}).
    \item \textbf{Mechanism:} We connect velocity scaling to population time lag and to earlier training and sampling corrections for state--time mismatch (Sections~\ref{sec:related} and~\ref{sec:catchup}).
          Our controls support lag reduction as an explanation for why velocity scaling improves image quality (Section~\ref{sec:response}).
    \item \textbf{Calibration:} We select the lag-based gain without using FID and show that it improves ImageNet FID.
          One fixed factor per model describes the additional gain beyond lag correction at higher NFEs, with a mean absolute gain error of 0.011 across four SiT sizes (Section~\ref{sec:gain-settings}).
\end{itemize}

\section{Related Work}%
\label{sec:related}

\subsection{Flow matching for image generation}

Generative flows learn a velocity field that transports a source distribution to the data distribution.
Likelihood-based continuous normalizing flows require numerically integrating the differential equation during training~\citep{chen2018neuralode}.
Flow Matching avoids this simulation by regressing the velocity field against target velocities from prescribed paths~\citep{lipman2023flowmatching}.
DiT introduced a scalable transformer backbone for diffusion models~\citep{peebles2023dit}.
SiT kept the same architecture but placed it in the stochastic-interpolant framework, which supports flow and diffusion objectives and deterministic or stochastic sampling~\citep{ma2024sit}.
REPA improved SiT training by aligning its hidden representations with features from pretrained visual encoders~\citep{yu2025repa}.
We study how inference-time velocity scaling changes the sampling dynamics of these models.

\subsection{State--time mismatch}

State--time mismatch occurs when a sampled state differs from the genuine states associated with its model time.
Prior work has addressed this mismatch through changes to training or sampling.

\paragraph{Training corrections.}
Input Perturbation assumes that Gaussian perturbations of diffusion training states can simulate errors accumulated earlier in sampling.
It trains the denoiser on these perturbed states so that it learns to recover from accumulated sampling errors~\citep{ning2023input}.
MixFlow observes that generated states lag behind their scheduled time, and trains the model at time $t$ on states from earlier times~\citep{li2025mixflow}.
DEFAR simulates one inference step during training to expose the model to a state affected by its own sampling error.
It trains the model to predict a velocity that moves this state toward the target image~\citep{huang2026defar}.
These methods change the training distribution and thereby bias the learned field toward a chosen model of sampling error.
Because the mismatch changes with NFE, solver, and guidance, a single training-time correction must cover several sampling settings or require further training.

\paragraph{Sampling corrections.}
Epsilon Scaling is a close diffusion analogue: it divides predicted noise by a factor above one.
This scaling is motivated by larger noise-prediction norms on generated than genuine states at the same time.
The factor is selected through an FID search~\citep{ning2024exposure}.
Time-Shift Sampler uses the variance across pixel and channel values within each generated image to infer its effective diffusion time, then supplies that time to the next model call~\citep{li2024timeshift}.

\paragraph{Removing noise-level conditioning.}
Another approach removes noise-level conditioning altogether, requiring the denoiser to infer corruption strength from the sampled state.
\citet{sun2025noiseconditioning} show that most diffusion, flow-matching, and consistency models remain functional without noise conditioning.
\citet{kadkhodaie2026blind} study the same blind formulation and connect it directly to state--time mismatch: the denoiser infers the current noise level from the state and follows an implicit schedule, adapting when residual corruption does not match the scheduled level.
Equivariant blind denoisers can even generalize to noise levels never seen during training without receiving the test noise level: \citet{mohan2020robust} demonstrate this with bias-free, scale-equivariant CNNs, and \citet{saied2026wne} extend the same robustness to arbitrary architectures.

Velocity scaling addresses the same state--time mismatch by advancing a lagging state toward its scheduled time (Section~\ref{sec:catchup}).
The lag fit selects a separate correction for each model and sampling setting without retraining, online time estimation, or an image metric.
Beyond correcting population time lag, we find that a well-chosen gain can improve sample quality in ways that lag correction alone does not explain (Section~\ref{sec:gain-settings}).

\section{Setup}%
\label{sec:setup}

\paragraph{Flow matching.}
For $t\in[0,1]$, let $X_0$ be source noise, $X_1$ be data, and $U=X_1-X_0$.
The linear training path runs from noise at $t=0$ to data at $t=1$:
\begin{equation}
    X_t=(1-t)X_0+tX_1.
    \label{eq:path}
\end{equation}
The population minimizer of the flow-matching loss is
\begin{equation}
    v^*(x,t)=\mathbb E[U\mid X_t=x].
    \label{eq:population-field}
\end{equation}
We distinguish this population field from the learned velocity field $v_\theta$ used during sampling.

\paragraph{Velocity scaling.}
During sampling, we multiply the model's velocity prediction by a scalar gain $\gamma(t)$ wherever that prediction enters the sampler:
\begin{equation}
    v_\theta^\gamma(x,t)=\gamma(t)v_\theta(x,t).
    \label{eq:velocity-scaling}
\end{equation}
Velocity scaling does not multiply the state or stochastic noise terms.
With $\gamma(t)=1$, the velocity is unchanged; we call this \emph{unscaled sampling}.
Constant gain means $\gamma(t)=c$ for all $t$.
\citet{li2026velocitydeficit} use $\gamma(t)=1.1-0.1t$ in their Scale Schedule Corrector (SSC).
In their SiT-XL/2 NFE-50 ablation, constant gain 1.05 gave FID 7.8 versus 7.6 for SSC, while several schedules with similar mean gain gave FID between 7.5 and 7.8.
These results suggest that schedule shape has little effect on FID.
We therefore use constant gain unless stated otherwise. 

\paragraph{Experimental scope.}
Under deterministic and stochastic sampling, we test velocity scaling across SiT model sizes, REPA-SiT, an ADM U-Net, and a decoder-free NCSN++ model trained on CIFAR-10.
Appendix~\ref{app:design} gives the experimental setup.

\section{No Velocity Deficit}%
\label{sec:no-deficit}

\paragraph{The population field has no velocity deficit.}
Whenever target velocity \(U\) has nonzero variance given state \(X_t\), Jensen's inequality implies that its conditional mean has a smaller squared norm than the target itself: \(\lVert \mathbb E[U\mid X_t]\rVert^2 < \mathbb E[\lVert U\rVert^2\mid X_t]\).
\citet{li2026velocitydeficit} interpret this gap as velocity magnitude lost to MSE training.
The standard variance identity reveals what this gap actually represents:
\begin{equation}
    \mathbb E[\lVert U\rVert^2\mid X_t]
    =\lVert v^*(X_t,t)\rVert^2
    +\mathbb E[\lVert U-v^*(X_t,t)\rVert^2\mid X_t].
    \label{eq:variance-identity}
\end{equation}
The second term measures conditional target variation.
It reflects valid training trajectories that cross at state \(X_t\), not missing velocity.
The conditional mean \(v^*\) remains the exact population velocity field that transports the source to the data distribution~\citep{lipman2023flowmatching}.
A Gaussian example makes this distinction concrete: conditional target variance is positive, yet gain one gives exact transport, while any other gain changes the endpoint covariance (Appendix~\ref{app:gaussian}).

\Needspace{15\baselineskip}
\paragraph{The trained field is not systematically underscaled.}
Although the population MSE solution is exact, finite training could leave the learned field systematically underscaled.
We tested this possibility at 20 values of \(t\) by fitting the scalar multiplier that minimizes held-out flow-matching error on 16,000 ImageNet image--noise pairs:
\begin{equation}
    \widehat\gamma_{\mathrm{FM}}(t)
    =\frac{\mathbb E[\langle v_\theta(X_t,t),U\rangle\mid t]}
    {\mathbb E[\lVert v_\theta(X_t,t)\rVert^2\mid t]}.
    \label{eq:fitted-model-gain}
\end{equation}
At fixed $t$, the expectations in Equation~\ref{eq:fitted-model-gain} are over independent draws of $X_0$ from the source distribution and $X_1$ from the data distribution.
Under this joint distribution, $v^*(X_t,t)=\mathbb E[U\mid X_t]$, so replacing $U$ with $v^*(X_t,t)$ leaves the numerator unchanged.
Thus $\widehat\gamma_{\mathrm{FM}}(t)$ is also the least-squares scalar that makes $\widehat\gamma_{\mathrm{FM}}(t)v_\theta$ best match the population field on genuine training-path states.
Its mean is 0.999, and the fitted curve remains near one at every time (Appendix Figure~\ref{fig:fitted-gamma}).
The tested SiT model is therefore not systematically underscaled on genuine training-path states.

Together, the population and finite-model results leave open why velocity scaling improves generation quality.
One possibility is that sampled states differ from genuine training-path states and that velocity scaling reduces this mismatch.
The next section examines this possibility.

\section{Population Time Lag}%
\label{sec:catchup}

We measure population time lag and compare two ways to address it: moving the model's time input backward and advancing sampled states through velocity scaling.
We show that choosing gain from the measured lag improves generation quality.

\subsection{Measuring Population Time Lag}%
\label{sec:lag-measurement}

Let $y$ denote the image class label.
At model time $t$, let $p_t(x,y)$ be the joint distribution of genuine path states and class labels, and $q_t^u(x,y)$ the corresponding distribution under unscaled sampling.
\citet{kadkhodaie2026blind} use a learned noise-level estimator to measure schedule mismatch in diffusion sampling.
Similarly, we train a probe \(\widehat s_\phi(x,y)\) on genuine path states to estimate path time from the state $x$ and class label $y$.
We define the population time lag measured by probe $\phi$ as
\begin{equation}
    \widehat\delta_\phi(t)
    =
    \mathbb E_{(x,y)\sim p_t}\!\left[\widehat s_\phi(x,y)\right]
    -
    \mathbb E_{(x,y)\sim q_t^u}\!\left[\widehat s_\phi(x,y)\right].
    \label{eq:lag}
\end{equation}
Subtracting the baseline expectation on genuine states ensures that intrinsic probe bias is not counted as sampling lag (Appendix~\ref{app:lag-validation}).
A positive value means that generated states resemble earlier points along the path than the current solver time $t$.
We use \emph{catch-up} to mean reducing this population time lag.

\paragraph{Measuring systematic lag.}
Even at a fixed true time, probe readings have high variance across genuine states (Appendix~\ref{app:lag-validation}).
This makes individual lag estimates unreliable, so we average across samples to detect systematic population time lag.

\paragraph{Validating the probe.}
We test whether the probe measures movement along a genuine path or simply responds to a change in the state.
Replacing each held-out genuine state with the state 0.05 time units earlier on the same path lowers the SiT probe's mean estimated time by 0.0499, almost exactly the known time change.
Moving the original states the same distance in a direction perpendicular to the path lowers the mean estimate by only 0.00519.
The probe therefore tracks the known change in path time and responds much less to perpendicular displacements (Appendix~\ref{app:lag-validation}).

\paragraph{Population time lag across models.}
We measure population time lag across the ImageNet models, but it is much smaller in the CIFAR-10 model (Appendix Table~\ref{tab:scalar-lag-calibration}).
Figure~\ref{fig:scalar-population-lag} shows how gain changes population time lag along sampled trajectories.

\Needspace{10\baselineskip}
\subsection{Model-Time Correction}%
\label{sec:model-time-correction}

Model-time correction replaces the time $t$ passed to the network with $t-\widehat\delta_\phi(t)$.
It leaves the integration clock unchanged and does not multiply the velocity.

\paragraph{Model-time correction improves FID.}
This model-time correction improves FID without scaling the velocity (at NFE 50, FID falls from 13.8 to 8.8; Appendix~\ref{app:model-time-correction}).

\paragraph{Model-time correction transfers to an ADM U-Net.}
The corresponding model-time correction lowers FID from 22.7 to 16.7 at NFE 50.
Constant gain 1.075 reaches FID 13.2, so measured timing affects image quality beyond SiT without accounting for the full scaling benefit (Appendix Table~\ref{tab:unet-time-correction}).

\subsection{Velocity Scaling and Population Time Lag}%
\label{sec:lag-correction}

Velocity scaling instead moves the state farther in the current model-velocity direction before the next model call.
Consider an Euler step of length $h$ from state $x_t$, with $v_t=v_\theta(x_t,t)$.
The solver advances to time $t+h$, but suppose the updated state resembles a genuine path state $x_\tau$ at an earlier effective time $t<\tau<t+h$:
\begin{equation}
    x_t+h v_t\approx x_\tau.
    \label{eq:ordinary-step}
\end{equation}
Under a local first-order approximation of the path, scaling the velocity gives
\begin{equation}
    x_t+\gamma h v_t
    \approx x_{t+\gamma(\tau-t)}
    \approx x_{t+h}
    \quad\text{when}\quad
    \gamma(\tau-t)\approx h.
    \label{eq:scaled-step}
\end{equation}
The next model call then receives a state that better matches its time input $t+h$.

\subsection{Selecting the Lag-Based Gain}%
\label{sec:lag-based-gain}

We select the \emph{lag-based gain}, $\gamma_{\rm lag}$, by searching for the constant gain that minimizes mean squared population time lag along the sampled trajectories.
We estimate this gain separately at each NFE for the four SiT sizes, the ADM U-Net, and CIFAR-10 NCSN++.
The objective uses neither image decoding nor an image-quality metric.
We can calibrate the gain at 0.01 resolution using 256 trajectories per candidate.
Appendix~\ref{app:scalar-calibration} gives the search procedure, cost, and stability checks.
Appendix~\ref{app:lag-gain-construction} compares scalar calibration with fitting a time-varying schedule.

\begin{figure}[ht]
    \centering
    \includegraphics[width=\textwidth]{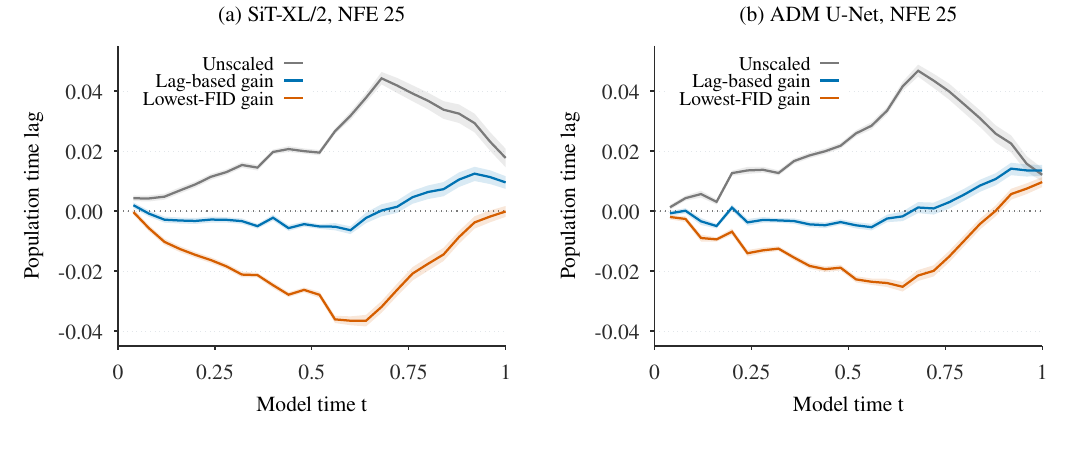}
    \caption{
        Population time lag under constant gains for unguided SiT-XL/2 and ADM U-Net at NFE 25.
        Positive lag means generated states read as earlier than genuine states; zero is the timing target.
        Each curve uses 1,280 paired paths, and shading gives pointwise 95\% whole-trajectory bootstrap intervals conditional on the chosen gain, trained probe, and genuine baseline.
        The same paths select the lag-based gain and display its lag.
        The lowest-FID gains come from separate 50,000-image gain grids.
    }%
    \label{fig:scalar-population-lag}
\end{figure}

\Needspace{7\baselineskip}
\paragraph{Lag-based gain improves generation quality.}
Direct scalar calibration reduces measured population time lag across all ImageNet models (Figure~\ref{fig:scalar-population-lag}; Appendix Table~\ref{tab:scalar-lag-validation}).
For all ImageNet models, lag-based gain selection also improves FID under coarse sampling.
Although selected without an image metric, these gains recover an estimated 49\% of the FID reduction achieved by the lowest-FID gain, averaged across the twenty ImageNet settings.
The estimate is 81\% at NFE 25 and falls with finer sampling (Appendix Table~\ref{tab:scalar-fid-recovery}).
The CIFAR-10 model has little population time lag, and its lowest-FID gain is one.

\FloatBarrier{}
\section{Changed-State Response}%
\label{sec:response}

When a state lags, the model is called with a time ahead of the state's effective time.
Velocity scaling moves the state forward, reducing this mismatch before subsequent model calls.
We use controls to test alternative explanations for the improvement from velocity scaling.

Consider two deterministic Euler trajectories with the same initial state and time grid: an unscaled trajectory with states $x_k$, and a scaled trajectory with states $y_k$.
Keep the class label fixed when the model is class conditional.
Their difference $\Delta_k=y_k-x_k$ obeys:
\begin{equation}
    \begin{aligned}
        x_{k+1} & =x_k+h_k v_\theta(x_k,t_k),                                       \\
        y_{k+1} & =y_k+h_k\gamma_k v_\theta(y_k,t_k),                               \\
        \Delta_{k+1}
                & =\Delta_k
        +\underbrace{h_k(\gamma_k-1)v_\theta(x_k,t_k)}_{\text{direct extra push}}     \\
                & \quad+h_k\gamma_k\underbrace{\bigl[v_\theta(y_k,t_k)-v_\theta(x_k,t_k)\bigr]}_{\text{changed-state response}}.
    \end{aligned}
    \label{eq:intuitive-path-decomposition}
\end{equation}
\Needspace{5\baselineskip}
The direct extra push is parallel to the unscaled velocity.
The changed-state response comes from evaluating the model at the moved state.

\paragraph{Frozen-field control.}
From the same initial state, the frozen-field control follows the velocity predictions recorded along the unscaled trajectory, multiplied by the gain used for full velocity scaling.
The model never evaluates the moved control states, excluding the changed-state response (Appendix~\ref{app:path-decomposition}).
Across five SiT checkpoints under stochastic sampling, this control achieves only 22.4\% to 24.9\% of the FID reduction from full velocity scaling, so most of the benefit requires re-evaluating the model at the moved states.
Under deterministic sampling, the control achieves a larger share, reaching 53.9\% for SiT and 59.0\% for an independently trained ADM U-Net~\citep{dao2023lfm} (Appendix Table~\ref{tab:response-transfer}).

\Needspace{7\baselineskip}
\paragraph{Radial control.}
Re-evaluating the model at moved states does not by itself guarantee better generation quality.
Let $x$ be the control's current state and $v=v_\theta(x,t)$ its predicted velocity.
The radial control replaces the extra velocity $(\gamma-1)v$ with $(\gamma-1)\lVert v\rVert x/\lVert x\rVert$, while still evaluating the model at the moved states.
This control worsens FID, whereas velocity scaling improves it (Appendix~\ref{app:direct-push-direction}).
The direction of the extra push therefore matters even when the model is re-evaluated throughout sampling.

Velocity scaling advances lagging states toward their scheduled time along the learned velocity (Section~\ref{sec:catchup}).
The frozen-field control shows that much of the improvement under stochastic sampling requires the model to predict new velocities at these advanced states.
The radial control shows that re-evaluation after a different displacement can instead worsen quality.
These findings support an explanation in which scaling improves generation by advancing lagging states and then using the learned field at states that better match its time input.

\section{Gain Across Settings}%
\label{sec:gain-settings}

We find that the lowest-FID gain is not an intrinsic property of the learned field: it shifts with numerical solver accuracy, guidance strength, and the chosen image metric.

\subsection{Numerical Accuracy and Gain}
\label{sec:numerical-accuracy}

\begin{figure}[!t]
    \centering
    \includegraphics[width=0.92\textwidth]{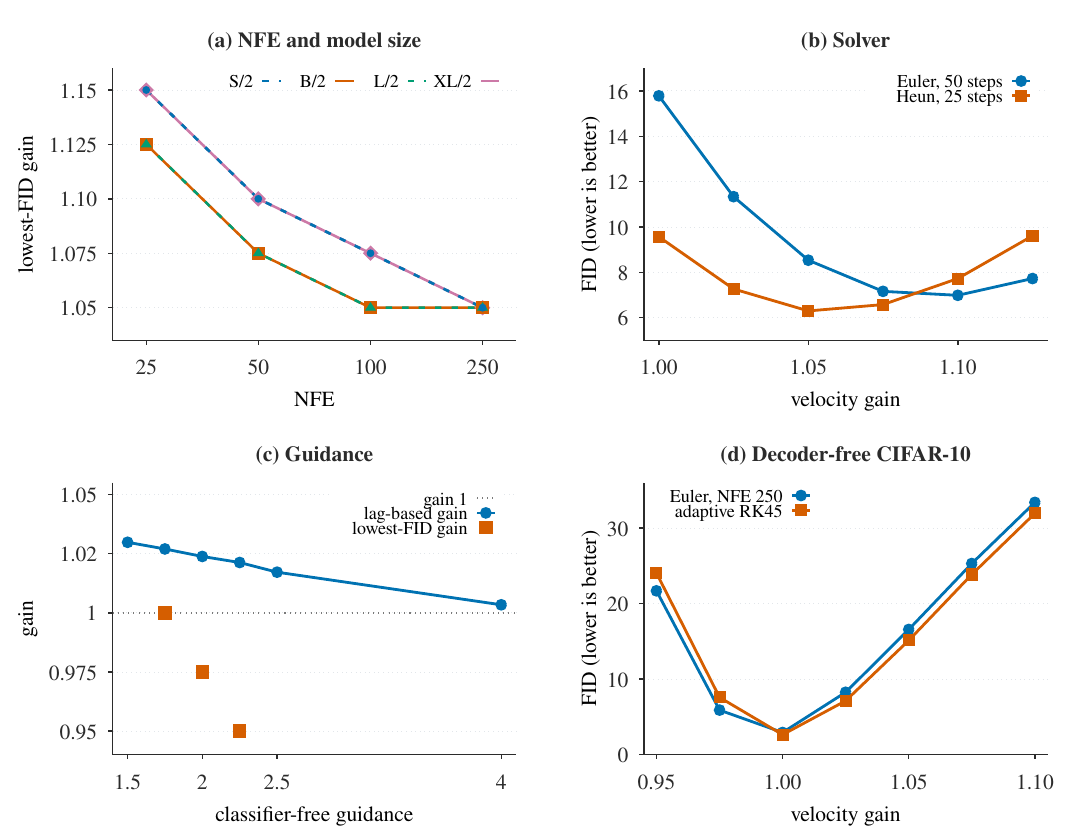}
    \caption{
        Gain selection changes with numerical accuracy, guidance, and sampling system.
        Panel (a) shows that the lowest-FID gain across four ImageNet SiT models moves toward one as NFE rises.
        Panel (b) shows that stochastic Heun selects a lower gain than Euler at a matched budget of 50 model evaluations.
        Panel (c) shows that the directly fitted lag-based gain and the lowest-FID gain fall as classifier-free guidance rises, but the lowest-FID gain falls faster.
        Each guided scalar fit uses 1,280 paired paths at NFE 50.
        Panel (d) shows that decoder-free CIFAR-10 selects gain one under fixed-step Euler and adaptive RK45.
        Each underlying FID evaluation uses 50,000 images per gain and the same random inputs across gains within each curve.
    }%
    \label{fig:context-dependence}
\end{figure}

More accurate integration reduces the gain favored by FID.
Across all ImageNet models, the lag-based gain approaches one as NFE increases (Appendix Table~\ref{tab:scalar-lag-calibration}).
The lowest-FID gain also moves toward one in all ImageNet models, while staying at one in the CIFAR-10 model (Figure~\ref{fig:context-dependence}a,d).
For SiT-XL/2 at fixed NFE, stochastic Heun reduces both population time lag and the lowest-FID gain relative to Euler (Figure~\ref{fig:context-dependence}b).
Although better integration needs less gain, velocity scaling is not simply a correction for integration error; it changes the sampling path (Section~\ref{sec:response}; Appendix Table~\ref{tab:solver-refinement}).

\paragraph{Why coarse sampling can lag.}
As an extreme example, consider Euler sampling with only two steps (NFE~2).
Let $x(t)$ follow the exact population field $v^*$ from noise $x_0$, and let $\mu_y$ be the class mean.
The first step replaces the integral to $t=0.5$ with a single velocity evaluation:
\[
    \begin{aligned}
        x(1/2)
        &=x_0+\int_0^{1/2}v^*(x(s),s,y)\,ds,\\
        \widetilde x_{1/2}
        &=x_0+\tfrac12v^*(x_0,0,y)
        =\tfrac12x_0+\tfrac12\mu_y.
    \end{aligned}
\]

\begin{samepage}
Intermediate velocity updates develop the image structure missing from the Euler states.
Comparing the Euler states with genuine states at $t=0.5$ tests whether the probe reads missing image structure as lag.
To focus this comparison on image structure, the probe receives normalized states during both training and evaluation.
With overall norm removed as a time cue, the probe assigns the Euler states earlier times on average than the genuine states (Appendix~\ref{app:coarse-refinement}).
The delayed and incomplete object development along the coarse unscaled path in Figure~\ref{fig:visual-clock} is consistent with this account of missing refinement.
\par
\end{samepage}

\Needspace{5\baselineskip}
\paragraph{Lag explains only part of the gain.}
Across all twenty tested unguided stochastic ImageNet settings, the lowest-FID gain exceeds the lag-based gain (Appendix Table~\ref{tab:scalar-lag-calibration}).
This gap persists with finer integration, even when measured population time lag is small.

\Needspace{9\baselineskip}
\subsection{Gain Beyond Lag Correction}%
\begin{figure}[!ht]
    \centering
    \includegraphics[width=\textwidth]{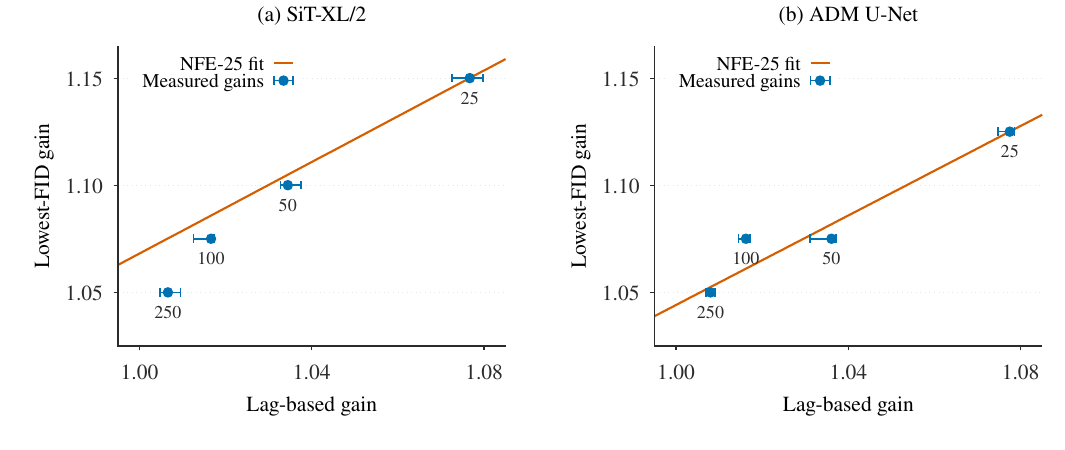}
    \caption{
        Direct scalar lag gains and lowest-FID gains for unguided ImageNet-256 SiT-XL/2 and ADM U-Net.
        Labels give NFE; horizontal bars are paired 95\% intervals selecting the best gain among measured candidates in 2,000 resamples of 1,280 paths, with the trained probe and genuine baseline fixed.
        Vertical coordinates are minima from existing 50,000-image FID grids, with no continuous FID-gain refinement.
        Each line tests whether the multiplier fitted at NFE 25 describes the gains at the other NFEs.
        Distances in gain do not measure FID differences.
        SiT uses uniform stochastic Euler and ADM uses Euler--Maruyama with a fixed final Mean step.
    }%
    \label{fig:scalar-gain-relation}
\end{figure}

\label{sec:gain-prediction}
We test whether one fixed factor per model describes the additional gain favored by FID beyond lag correction across NFEs.
For each ImageNet model, let $\gamma^\star_{\rm FID}(N)$ be the lowest-FID gain at NFE $N$.
This hypothesis gives
\begin{equation}
    \gamma^\star_{\rm FID}(N)
    \approx\kappa\gamma_{\rm lag}(N),
    \qquad
    \kappa=\frac{\gamma^\star_{\rm FID}(25)}{\gamma_{\rm lag}(25)}.
    \label{eq:additional-gain}
\end{equation}
One fixed multiplier per model roughly relates the lag-based gain to the lowest-FID gain across NFEs (Figure~\ref{fig:scalar-gain-relation}).

At NFE 250, the $\kappa$-adjusted gains give FIDs close to the lowest measured FID for every ImageNet model (Table~\ref{tab:gain-fid-summary}).
\begin{table}[!ht]
\centering
\small
\setlength{\tabcolsep}{5pt}
\caption{
ImageNet-256 FID at NFE 25 and 250; lower is better.
Lag-gain FID is linearly interpolated between the measured gains bracketing the unrounded $\gamma_{\rm lag}(N)$.
The $\kappa$-adjusted column gives measured FID at the closest tested gain to $\kappa\gamma_{\rm lag}(N)$, including gain one.
Each model's $\kappa$ is fitted at NFE 25 and reused at NFE 250.
Each measured FID uses 50,000 paired images within its model and NFE; these summaries have no computed uncertainty intervals.
SiT uses uniform stochastic Euler and ADM uses Euler--Maruyama with a final Mean step.
}
\label{tab:gain-fid-summary}
\begin{tabular}{@{}lrrrrr@{}}
\toprule
Model & NFE & \shortstack{Unscaled\\FID} & \shortstack{Lag-gain FID\\(interpolated)} & \shortstack{$\kappa$-adjusted\\FID} & \shortstack{Lowest measured\\FID} \\
\midrule
SiT-S/2 & 25 & 71.1 & 50.9 & 44.4 & 44.4 \\
 & 250 & 50.6 & 49.9 & 44.1 & 43.8 \\
SiT-B/2 & 25 & 47.9 & 30.2 & 25.9 & 25.9 \\
 & 250 & 29.1 & 28.5 & 24.8 & 24.8 \\
SiT-L/2 & 25 & 30.2 & 16.1 & 13.4 & 13.4 \\
 & 250 & 14.6 & 14.4 & 12.2 & 12.2 \\
SiT-XL/2 & 25 & 28.0 & 12.2 & 8.7 & 8.7 \\
 & 250 & 9.7 & 9.1 & 6.6 & 6.3 \\
ADM U-Net & 25 & 34.8 & 17.9 & 14.7 & 14.7 \\
 & 250 & 16.8 & 15.8 & 12.8 & 12.8 \\
\bottomrule
\end{tabular}
\end{table}

\begin{figure}[!ht]
    \centering
    \includegraphics[width=0.85\textwidth]{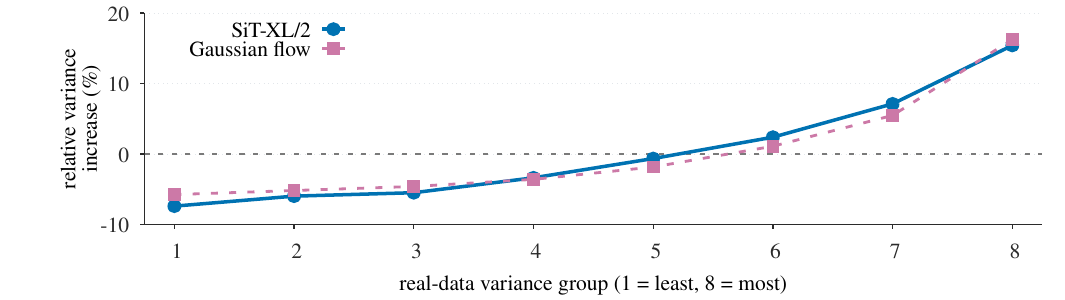}
    \caption{
        Scaling gives larger proportional variance increases to directions with larger real-data variance.
        We compare gain 1.05 with unscaled sampling for unguided SiT-XL/2 and a matched Gaussian flow under stochastic Euler sampling at NFE 250.
        Directions are grouped by held-out real-data variance.
        Each point shows the geometric-mean scaled-to-unscaled variance ratio within a group, divided by that across all directions and expressed as a percentage change from one.
        Negative values indicate a smaller proportional increase than the overall increase.
        SiT uses 1,280 paired endpoints per gain and 1,024 real-data latents; Appendix~\ref{app:spectral-sharpening} gives the covariance-slope measurements across NFEs and their uncertainty.
    }%
    \label{fig:spectral-redistribution}
\end{figure}

\FloatBarrier{}
\Needspace{9\baselineskip}
\paragraph{Scaling sharpens endpoint covariance.}
Velocity scaling also changes how endpoint variance is distributed: it gives larger proportional increases to directions in which real data vary more.
The endpoint distribution generated by SiT shows the same kind of covariance sharpening as in the exact Gaussian example (Figure~\ref{fig:spectral-redistribution}; Appendix~\ref{app:spectral-sharpening}).
We observe this sharpening at both low and high NFE.

To relate this sharpening to $\kappa$, we start with the best tested NFE-25 gain for unguided SiT-XL/2.
We measure how much it sharpens the endpoint covariance compared with the lag-based gain.
At each higher NFE, we find the gain that gives the same relative sharpening.
The gains we find are close to those predicted by $\kappa$.
Their estimated FIDs are within 0.2 of the lowest measured values at NFE 50, 100, and 250 (Appendix Figure~\ref{fig:sharpening-transfer}).

\FloatBarrier{}
\Needspace{10\baselineskip}
\subsection{Scope of Gain Selection}

\paragraph{Guidance.}

Classifier-free guidance combines predictions made with and without the class label to control how strongly the label affects sampling~\citep{ho2022classifierfree}.
As guidance strength \(\omega\) increases, the lowest-FID gain decreases and crosses below one (Figure~\ref{fig:context-dependence}c).
Stronger guidance also lowers the lag-based gain, but the lowest-FID gain falls faster (Appendix~\ref{app:cfg-gain}).
Guidance also increases velocity magnitude, but the measured increase is too small to explain the full decrease in the lowest-FID gain (Appendix~\ref{app:cfg-gain}).

\paragraph{Image metrics.}

Even within one sampler, different image metrics can favor different gains.
For SiT-XL/2 at NFE 25, FID favors a higher gain than sFID.
That higher gain also raises precision and lowers recall (Appendix Table~\ref{tab:nfe25-metric-tradeoff}).

\Needspace{8\baselineskip}
\paragraph{Dataset.}

The decoder-free NCSN++ model trained on CIFAR-10 provides a setting where velocity scaling is unnecessary~\citep{liu2022flow}.
Across deterministic Euler, adaptive RK45, and pixel-space Euler--Maruyama curves, the lowest measured FID occurs at gain one.
The lag-based gain stays close to one at all four NFEs (Appendix~\ref{app:scalar-lag-calibration}).
This result matches the catch-up account: population time lag is small and gain above one provides no benefit (Figure~\ref{fig:context-dependence}d).

\begin{samepage}
\section{Conclusion}

Mean-squared-error training does not create a velocity deficit: the population minimizer is exact, and the learned field shows no systematic norm bias.
In coarse samplers, sampled states fall behind their model time, exhibiting population time lag.
Velocity scaling advances these lagging states and changes subsequent velocity predictions.
Under stochastic sampling, this changed-state response produces most of the FID reduction.
Across the tested NFEs, the additional gain favored by FID remains roughly proportional to the lag-based gain.
The lowest-FID gain changes with solver, guidance, and image metric, reflecting the sampling and evaluation setting rather than the learned field alone.
\par
\end{samepage}

\section*{Reproducibility Statement}

The appendix specifies the models, data, samplers, gain grids, probe architecture and training, calibration cost and convergence, pairing, metrics, and uncertainty calculations.
Code will be released soon.

\Needspace{16\baselineskip}
\section*{AI Use Statement}

We used generative AI tools to help develop hypotheses and interpret experimental results.
The tools also assisted with experiment design and mathematical derivations.
They helped implement and debug the experimental and analysis code.
We used them to discuss related work and to draft and revise the manuscript and figure code.

The authors directed the research and assessed the scientific claims.
We checked derivations with analytical and numerical controls, and checked code with focused tests and reference comparisons.
The authors take responsibility for the final text, claims, and results.

\bibliography{references}

\begin{thebibliography}{29}
\providecommand{\natexlab}[1]{#1}
\providecommand{\url}[1]{\texttt{#1}}
\expandafter\ifx\csname urlstyle\endcsname\relax
  \providecommand{\doi}[1]{doi: #1}\else
  \providecommand{\doi}{doi: \begingroup \urlstyle{rm}\Url}\fi

\bibitem[Brent(1973)]{brent1973minimization}
Richard~P. Brent.
\newblock \emph{Algorithms for Minimization without Derivatives}.
\newblock Prentice-Hall, Englewood Cliffs, New Jersey, 1973.

\bibitem[Chen et~al.(2018)Chen, Rubanova, Bettencourt, and
  Duvenaud]{chen2018neuralode}
Ricky T.~Q. Chen, Yulia Rubanova, Jesse Bettencourt, and David Duvenaud.
\newblock Neural ordinary differential equations.
\newblock In \emph{Advances in Neural Information Processing Systems}, 2018.

\bibitem[Dao et~al.(2023)Dao, Phung, Nguyen, and Tran]{dao2023lfm}
Quan Dao, Hao Phung, Binh Nguyen, and Anh Tran.
\newblock Flow matching in latent space.
\newblock \emph{arXiv preprint arXiv:2307.08698}, 2023.
\newblock \doi{10.48550/arXiv.2307.08698}.

\bibitem[Deng et~al.(2009)Deng, Dong, Socher, Li, Li, and
  Fei-Fei]{deng2009imagenet}
Jia Deng, Wei Dong, Richard Socher, Li-Jia Li, Kai Li, and Li~Fei-Fei.
\newblock {ImageNet}: A large-scale hierarchical image database.
\newblock In \emph{IEEE Conference on Computer Vision and Pattern Recognition},
  pp.\  248--255, 2009.
\newblock \doi{10.1109/CVPR.2009.5206848}.

\bibitem[Dhariwal \& Nichol(2021)Dhariwal and Nichol]{dhariwal2021adm}
Prafulla Dhariwal and Alexander Nichol.
\newblock Diffusion models beat {GANs} on image synthesis.
\newblock In \emph{Advances in Neural Information Processing Systems}, 2021.

\bibitem[Heusel et~al.(2017)Heusel, Ramsauer, Unterthiner, Nessler, and
  Hochreiter]{heusel2017fid}
Martin Heusel, Hubert Ramsauer, Thomas Unterthiner, Bernhard Nessler, and Sepp
  Hochreiter.
\newblock {GANs} trained by a two time-scale update rule converge to a local
  nash equilibrium.
\newblock In \emph{Advances in Neural Information Processing Systems}, 2017.

\bibitem[Ho \& Salimans(2021)Ho and Salimans]{ho2022classifierfree}
Jonathan Ho and Tim Salimans.
\newblock Classifier-free diffusion guidance.
\newblock In \emph{NeurIPS Workshop on Deep Generative Models and Downstream
  Applications}, 2021.
\newblock URL \url{https://openreview.net/forum?id=qw8AKxfYbI}.

\bibitem[Huang et~al.(2026)Huang, Mao, Huang, Liu, Luo, Liang, Lu, Wang, Liu,
  Fu, Huang, and Huang]{huang2026defar}
Guanbo Huang, Jingjia Mao, Fanding Huang, Fengkai Liu, Xiangyang Luo, Yaoyuan
  Liang, Jiasheng Lu, Xiaoe Wang, Pei Liu, Ruiliu Fu, Ruqi Huang, and Shao-Lun
  Huang.
\newblock Exposure bias can alleviate itself via directional and frequency
  rectification in flow matching.
\newblock In \emph{European Conference on Computer Vision}, 2026.

\bibitem[Kadkhodaie et~al.(2026)Kadkhodaie, Pooladian, Chewi, and
  Simoncelli]{kadkhodaie2026blind}
Zahra Kadkhodaie, Aram-Alexandre Pooladian, Sinho Chewi, and Eero~P.
  Simoncelli.
\newblock Blind denoising diffusion models and the blessings of dimensionality.
\newblock \emph{arXiv preprint arXiv:2602.09639}, 2026.

\bibitem[Krizhevsky(2009)]{krizhevsky2009cifar}
Alex Krizhevsky.
\newblock Learning multiple layers of features from tiny images.
\newblock Technical report, University of Toronto, 2009.
\newblock URL
  \url{https://www.cs.toronto.edu/~kriz/learning-features-2009-TR.pdf}.

\bibitem[Kynk{\"a}{\"a}nniemi et~al.(2019)Kynk{\"a}{\"a}nniemi, Karras, Laine,
  Lehtinen, and Aila]{kynkaanniemi2019precision}
Tuomas Kynk{\"a}{\"a}nniemi, Tero Karras, Samuli Laine, Jaakko Lehtinen, and
  Timo Aila.
\newblock Improved precision and recall metric for assessing generative models.
\newblock In \emph{Advances in Neural Information Processing Systems}, 2019.

\bibitem[Li et~al.(2026{\natexlab{a}})Li, Lyu, Wang, Cheng, Zhu, and
  Wang]{li2025mixflow}
Hui Li, Jiayue Lyu, Fu-Yun Wang, Kaihui Cheng, Siyu Zhu, and Jingdong Wang.
\newblock {MixFlow} training: Alleviating exposure bias with slowed
  interpolation mixture.
\newblock In \emph{IEEE/CVF Conference on Computer Vision and Pattern
  Recognition}, pp.\  9095--9105, 2026{\natexlab{a}}.

\bibitem[Li et~al.(2026{\natexlab{b}})Li, Hong, Zhang, Lin, Li, Tang, and
  Liang]{li2026velocitydeficit}
Linze Li, Zong-Wei Hong, Shen Zhang, Bo~Lin, Jinglun Li, Yao Tang, and Jiajun
  Liang.
\newblock The velocity deficit: Initial energy injection for flow matching.
\newblock In \emph{International Conference on Machine Learning},
  2026{\natexlab{b}}.
\newblock URL \url{https://openreview.net/forum?id=ddBkQ4koaI}.

\bibitem[Li et~al.(2024)Li, Qu, Yao, Sun, and Moens]{li2024timeshift}
Mingxiao Li, Tingyu Qu, Ruicong Yao, Wei Sun, and Marie-Francine Moens.
\newblock Alleviating exposure bias in diffusion models through sampling with
  shifted time steps.
\newblock In \emph{International Conference on Learning Representations}, 2024.

\bibitem[Lipman et~al.(2023)Lipman, Chen, Ben-Hamu, Nickel, and
  Le]{lipman2023flowmatching}
Yaron Lipman, Ricky T.~Q. Chen, Heli Ben-Hamu, Maximilian Nickel, and Matthew
  Le.
\newblock Flow matching for generative modeling.
\newblock In \emph{International Conference on Learning Representations}, 2023.
\newblock URL \url{https://openreview.net/forum?id=PqvMRDCJT9t}.

\bibitem[Liu et~al.(2023)Liu, Gong, and Liu]{liu2022flow}
Xingchao Liu, Chengyue Gong, and Qiang Liu.
\newblock Flow straight and fast: Learning to generate and transfer data with
  rectified flow.
\newblock In \emph{International Conference on Learning Representations}, 2023.
\newblock URL \url{https://openreview.net/forum?id=XVjTT1nw5z}.

\bibitem[Loshchilov \& Hutter(2019)Loshchilov and Hutter]{loshchilov2019adamw}
Ilya Loshchilov and Frank Hutter.
\newblock Decoupled weight decay regularization.
\newblock In \emph{International Conference on Learning Representations}, 2019.
\newblock URL \url{https://openreview.net/forum?id=Bkg6RiCqY7}.

\bibitem[Ma et~al.(2024)Ma, Goldstein, Albergo, Boffi, Vanden-Eijnden, and
  Xie]{ma2024sit}
Nanye Ma, Mark Goldstein, Michael~S. Albergo, Nicholas~M. Boffi, Eric
  Vanden-Eijnden, and Saining Xie.
\newblock {SiT}: Exploring flow and diffusion-based generative models with
  scalable interpolant transformers.
\newblock In \emph{European Conference on Computer Vision}, pp.\  23--40, 2024.
\newblock \doi{10.1007/978-3-031-72980-5_2}.

\bibitem[Mohan et~al.(2020)Mohan, Kadkhodaie, Simoncelli, and
  Fernandez-Granda]{mohan2020robust}
Sreyas Mohan, Zahra Kadkhodaie, Eero~P. Simoncelli, and Carlos
  Fernandez-Granda.
\newblock Robust and interpretable blind image denoising via bias-free
  convolutional neural networks.
\newblock In \emph{International Conference on Learning Representations}, 2020.

\bibitem[Naeem et~al.(2020)Naeem, Oh, Uh, Choi, and Yoo]{naeem2020density}
Muhammad~Ferjad Naeem, Seong~Joon Oh, Youngjung Uh, Yunjey Choi, and Jaejun
  Yoo.
\newblock Reliable fidelity and diversity metrics for generative models.
\newblock In \emph{International Conference on Machine Learning}, pp.\
  7176--7185, 2020.

\bibitem[Nash et~al.(2021)Nash, Menick, Dieleman, and Battaglia]{nash2021sfid}
Charlie Nash, Jacob Menick, Sander Dieleman, and Peter Battaglia.
\newblock Generating images with sparse representations.
\newblock In \emph{International Conference on Machine Learning}, pp.\
  7958--7968, 2021.

\bibitem[Ning et~al.(2023)Ning, Sangineto, Porrello, Calderara, and
  Cucchiara]{ning2023input}
Mang Ning, Enver Sangineto, Angelo Porrello, Simone Calderara, and Rita
  Cucchiara.
\newblock Input perturbation reduces exposure bias in diffusion models.
\newblock In \emph{International Conference on Machine Learning}, pp.\
  26245--26265, 2023.

\bibitem[Ning et~al.(2024)Ning, Li, Su, Salah, and Ertugrul]{ning2024exposure}
Mang Ning, Mingxiao Li, Jianlin Su, Albert~Ali Salah, and Itir~Onal Ertugrul.
\newblock Elucidating the exposure bias in diffusion models.
\newblock In \emph{International Conference on Learning Representations}, 2024.

\bibitem[Oquab et~al.(2024)Oquab, Darcet, Moutakanni, Vo, Szafraniec, Khalidov,
  Fernandez, Haziza, Massa, El-Nouby, Assran, Ballas, Galuba, Howes, Huang, Li,
  Misra, Rabbat, Sharma, Synnaeve, Xu, J{\'e}gou, Mairal, Labatut, Joulin, and
  Bojanowski]{oquab2024dinov2}
Maxime Oquab, Timoth{\'e}e Darcet, Th{\'e}o Moutakanni, Huy Vo, Marc
  Szafraniec, Vasil Khalidov, Pierre Fernandez, Daniel Haziza, Francisco Massa,
  Alaaeldin El-Nouby, Mahmoud Assran, Nicolas Ballas, Wojciech Galuba, Russell
  Howes, Po-Yao Huang, Shang-Wen Li, Ishan Misra, Michael Rabbat, Vasu Sharma,
  Gabriel Synnaeve, Hu~Xu, Herv{\'e} J{\'e}gou, Julien Mairal, Patrick Labatut,
  Armand Joulin, and Piotr Bojanowski.
\newblock {DINOv2}: Learning robust visual features without supervision.
\newblock \emph{Transactions on Machine Learning Research}, 2024.

\bibitem[Peebles \& Xie(2023)Peebles and Xie]{peebles2023dit}
William Peebles and Saining Xie.
\newblock Scalable diffusion models with transformers.
\newblock In \emph{IEEE/CVF International Conference on Computer Vision}, pp.\
  4195--4205, 2023.
\newblock \doi{10.1109/ICCV51070.2023.00387}.

\bibitem[Saied \& Fleuret(2026)Saied and Fleuret]{saied2026wne}
Youssef Saied and Fran{\c{c}}ois Fleuret.
\newblock Normalization equivariance for arbitrary backbones, with application
  to image denoising.
\newblock In \emph{Proceedings of the 43rd International Conference on Machine
  Learning}, volume 306 of \emph{Proceedings of Machine Learning Research}.
  PMLR, 2026.

\bibitem[Salimans et~al.(2016)Salimans, Goodfellow, Zaremba, Cheung, Radford,
  and Chen]{salimans2016inceptionscore}
Tim Salimans, Ian Goodfellow, Wojciech Zaremba, Vicki Cheung, Alec Radford, and
  Xi~Chen.
\newblock Improved techniques for training {GANs}.
\newblock In \emph{Advances in Neural Information Processing Systems}, 2016.

\bibitem[Sun et~al.(2025)Sun, Jiang, Zhao, and He]{sun2025noiseconditioning}
Qiao Sun, Zhicheng Jiang, Hanhong Zhao, and Kaiming He.
\newblock Is noise conditioning necessary for denoising generative models?
\newblock In \emph{International Conference on Machine Learning}, pp.\
  57469--57502, 2025.

\bibitem[Yu et~al.(2025)Yu, Kwak, Jang, Jeong, Huang, Shin, and
  Xie]{yu2025repa}
Sihyun Yu, Sangkyung Kwak, Huiwon Jang, Jongheon Jeong, Jonathan Huang, Jinwoo
  Shin, and Saining Xie.
\newblock Representation alignment for generation: Training diffusion
  transformers is easier than you think.
\newblock In \emph{International Conference on Learning Representations}, 2025.

\end{thebibliography}
\bibliographystyle{iclr2027_conference}

\clearpage
\appendix

\section{Questions and main findings}%
\label{app:questions-findings}

\begin{table}[!ht]
    \centering
    \small
    \setlength{\tabcolsep}{4pt}
    \renewcommand{\arraystretch}{1.12}
    \caption{Questions, main findings, and supporting sections.}
    \label{tab:questions-findings}
    \begin{tabularx}{\textwidth}{@{}L{0.22\textwidth}>{\raggedright\arraybackslash}XL{0.17\textwidth}@{}}
        \toprule
        Question & Main finding & Where \\
        \midrule
        \textbf{Is scaling repairing underestimated velocities?} & The population MSE field gives exact transport. The tested pretrained SiT field has an MSE-optimal scalar multiplier near one on genuine training-path states. & \S\ref{sec:no-deficit}; Appendices~\ref{app:population}, \ref{app:trained-field-scaling} \\
        \addlinespace[0.7em]
        \textbf{Why can coarse sampling produce lag?} & Coarse integration can miss intermediate refinement. The two-step Euler example shows how missing sample-specific image structure registers as lag. & \S\ref{sec:numerical-accuracy}; Appendix~\ref{app:coarse-refinement} \\
        \addlinespace[0.7em]
        \textbf{Can we measure this lag?} & Comparing mean probe readings on generated and genuine states reveals systematic population time lag, which persists after global scale normalization. & \S\ref{sec:lag-measurement}; Appendix~\ref{app:state-time-probes} \\
        \addlinespace[0.7em]
        \textbf{How can scaling help states catch up?} & A first-order argument explains how scaling advances lagging states toward their scheduled time; moving model time backward addresses the same mismatch. Controls show that much of the benefit requires the model to predict new velocities at the advanced states. & \S\S\ref{sec:model-time-correction}--\ref{sec:lag-correction}, \ref{sec:response}; Appendices \mbox{\ref{app:model-time-correction}--\ref{app:unet-model-time-correction}}, \ref{app:response} \\
        \addlinespace[0.7em]
        \textbf{Can lag select a useful gain?} & Minimizing measured lag selects gains without an image-quality metric and improves coarse-sampling FID across the tested ImageNet models. The CIFAR-10 model has little lag and favors gain one. & \S\ref{sec:lag-based-gain}; Appendix~\ref{app:scalar-calibration} \\
        \addlinespace[0.7em]
        \textbf{What determines the useful gain beyond lag correction?} & The useful gain depends on solver accuracy, guidance, and the image metric. In unguided ImageNet experiments, one multiplier per model roughly relates lowest-FID gains to lag-based gains across NFEs. For unguided SiT-XL/2, sharpening predicts gains with estimated FID within 0.2 of the best measured value. & \S\ref{sec:gain-settings}; Appendices~\ref{app:gain-fid-models}, \ref{app:gain}--\ref{app:transfer} \\
        \bottomrule
    \end{tabularx}
\end{table}
\clearpage

\section{Theoretical analysis}%
\label{app:population}

\subsection{Residual target variation and exact transport}%
\label{app:residual-transport}

Flow matching assigns a target velocity $U$ to each sampled training path.
At a fixed state $X_t$ and time $t$, $U$ can still vary.
Different training paths can reach the same state.
The population MSE field is the mean target velocity at that state and time:
\begin{equation}
    v^*(X_t,t)=\mathbb E[U\mid X_t,t].
\end{equation}
Define the residual target variation by
\begin{equation}
    \varepsilon_t=U-v^*(X_t,t),
    \qquad
    \mathbb E[\varepsilon_t\mid X_t,t]=0.
\end{equation}
The residual $\varepsilon_t$ is not noise added to the model.
It represents the variation in $U$ that $X_t$ and $t$ do not determine.
Using $U=v^*(X_t,t)+\varepsilon_t$ and expanding the squared norm gives Equation~\ref{eq:variance-identity}.
The cross term has conditional mean zero.
The residual variance therefore equals the gap between the average squared target norm and the squared norm of $v^*$.
\begin{equation}
    \mathbb E[\lVert \varepsilon_t\rVert^2]
    =\mathbb E[\lVert U\rVert^2]
    -\mathbb E[\lVert v^*(X_t,t)\rVert^2].
\end{equation}
It is target variation, not a missing velocity component.

The continuity equation shows why this residual is not a transport error.
For any smooth test function $\varphi$ with bounded support,
\begin{align}
    \frac{d}{dt}\mathbb E[\varphi(X_t)]
     & =\mathbb E[{\nabla\varphi(X_t)}^\top U]                           \\
     & =\mathbb E[{\nabla\varphi(X_t)}^\top\{v^*(X_t,t)+\varepsilon_t\}] \\
     & =\mathbb E[{\nabla\varphi(X_t)}^\top v^*(X_t,t)].
\end{align}
The residual contributes zero because its conditional mean is zero.
When the continuity equation has a unique solution, $v^*$ transports the training distribution exactly at every time~\citep{lipman2023flowmatching}.
This residual variation is not a missing velocity component.

\subsection{Gaussian example: scaling changes exact transport}%
\label{app:gaussian}

A Gaussian path makes the general result concrete.
Its residual variance is positive.
Despite this residual variance, gain one gives exact transport.
Scaling changes the endpoint distribution.

Let $X_0\sim\mathcal N(0,I)$ and $X_1\sim\mathcal N(0,\Sigma)$ be independent, where $\Sigma$ is positive definite.
Define the linear path $X_t=(1-t)X_0+tX_1$ and its target velocity $U=X_1-X_0$.
The path covariance is
\begin{equation}
    Q_t={(1-t)}^2I+t^2\Sigma.
\end{equation}
Gaussian conditioning gives the population velocity
\begin{equation}
    v^*(x,t)
    =\left[t\Sigma-(1-t)I\right]Q_t^{-1}x
    =\frac{1}{2}\dot Q_t Q_t^{-1}x.
\end{equation}
The conditional residual covariance is
\begin{equation}
    \operatorname{Cov}(U\mid X_t)
    =I+\Sigma-\left[t\Sigma-(1-t)I\right]Q_t^{-1}
    \left[t\Sigma-(1-t)I\right]
    =\Sigma Q_t^{-1}.
\end{equation}
This covariance is positive definite.
The target--field norm gap is therefore strict.

Now multiply this exact field by a constant gain $\gamma$.
If $Y_\gamma$ solves $\dot Y_\gamma=\gamma v^*(Y_\gamma,t)$ with $Y_\gamma(0)\sim\mathcal N(0,I)$, then
\begin{equation}
    Y_\gamma(t)=Q_t^{\gamma/2}Y_\gamma(0).
\end{equation}
At the endpoint,
\begin{equation}
    Y_\gamma(1)=\Sigma^{\gamma/2}Y_\gamma(0),
    \qquad
    Y_\gamma(1)\sim\mathcal N(0,\Sigma^\gamma).
\end{equation}
At gain one, $Y_1(t)\sim\mathcal N(0,Q_t)$, so the field follows the full training path and reaches the target covariance $\Sigma$.
For $\gamma\ne1$, the endpoint covariance is instead $\Sigma^\gamma$ and generally differs from the target.
The gain changes each covariance eigenvalue from $\lambda$ to $\lambda^\gamma$.
For $\gamma>1$, directions with $\lambda>1$ expand and directions with $\lambda<1$ contract.
Scaling therefore changes an already correct transport map.
It does not restore the residual target variation.
The later experiments measure how that change appears in finite neural samplers.

\subsection{Magnitude-aware training changes the population target}%
\label{app:magnitude-objective}

Li et al.\ also propose a magnitude-aware loss that increases predicted velocity norms~\citep{li2026velocitydeficit}.
We ask whether this loss recovers the residual target variation from Appendix~\ref{app:residual-transport}.
Fix a state $x$ and time $t$, and write $\lambda=\lambda(t)$.
The conditional population objective is
\begin{equation}
    \mathcal L_\lambda(v\mid x,t)
    =
    \mathbb E\!\left[
        \lVert v-U\rVert^2
        +\lambda{\bigl(\lVert v\rVert-\lVert U\rVert\bigr)}^2
        \,\middle|\, X_t=x
        \right].
\end{equation}
Let $\mu=\mathbb E[U\mid X_t=x]$ and $m=\mathbb E[\lVert U\rVert\mid X_t=x]$.
For nonzero $\mu$, the population minimizer is
\begin{equation}
    v^*_{\lambda}(x,t)
    =\frac{\lVert\mu\rVert+\lambda m}{(1+\lambda)\lVert\mu\rVert}\mu.
\end{equation}
When $m>\lVert\mu\rVert$, the multiplier exceeds one.
The loss inflates the conditional mean instead of recovering the residual.
When $\mu=0$ and $m>0$, the magnitude term asks for nonzero norm but supplies no direction.
Magnitude-aware training therefore changes the population target.
Its effect on image quality is a separate question.

\section{Experimental setup}%
\label{app:design}

\subsection{Models}

\paragraph{SiT.}
The main experiments use public pretrained class-conditional SiT checkpoints~\citep{ma2024sit}.
The central experiments use SiT-XL/2.
The model-transfer experiments use SiT-S/2, B/2, L/2, and XL/2.
We also test a REPA-trained SiT-XL/2 checkpoint~\citep{yu2025repa}.

\paragraph{ADM U-Net.}
The architecture-transfer experiments use the released class-conditional latent flow-matching ADM U-Net~\citep{dhariwal2021adm,dao2023lfm}.
We decode its outputs with the matching variational autoencoder.

\paragraph{Rectified Flow.}
The decoder-free experiments use an NCSN++ model with the official pretrained 1-Rectified Flow checkpoint~\citep{liu2022flow}.
This model predicts centered pixels directly and does not use a variational autoencoder.

\subsection{Data}

\paragraph{ImageNet-256.}
The SiT and ADM U-Net experiments use class-conditional ImageNet-256~\citep{deng2009imagenet}.
Both model families sample $4\times32\times32$ latent states with 4,096 coordinates.
Their variational autoencoders decode these states into $3\times256\times256$ RGB images.
The latent scaling factor is 0.18215.
Generated labels are sampled uniformly from the 1,000 ImageNet classes.

The real states for the ImageNet probes use images from the training split.
We convert each image to RGB, resize and center-crop it to $256\times256$, and scale its pixels to $[-1,1]$.
We then sample the variational-autoencoder posterior and apply the same latent scaling factor.

\paragraph{CIFAR-10.}
The decoder-free experiments use CIFAR-10~\citep{krizhevsky2009cifar}.
Each model state contains centered $3\times32\times32$ RGB pixels.
This gives 3,072 coordinates and requires no encoder or decoder.
The CIFAR-10 probe uses randomly flipped images from the training split.

\paragraph{Counts and references.}
Table~\ref{tab:data-summary} gives the representation and sample count for each data use.
Unless stated otherwise, every generated-image setting contains 50,000 samples.
Saved trajectory and endpoint analyses use smaller sets, whose captions state their sample counts.

\begin{table}[ht]
    \centering
    \small
    \setlength{\tabcolsep}{4pt}
    \caption{
        Data representations, sample counts, and reference data.
        ImageNet probe counts apply separately to the SiT and ADM U-Net probes.
        Probability calibration uses genuine states that are separate from probe training and held-out evaluation.
    }%
    \label{tab:data-summary}
    \begin{tabularx}{\textwidth}{@{}L{3.0cm}L{3.4cm}L{3.4cm}>{\raggedright\arraybackslash}X@{}}
        \toprule
        Data use            & Representation                                                            & Count                                               & Source or reference                            \\
        \midrule
        ImageNet generation & $3\times256\times256$ RGB image decoded from a $4\times32\times32$ latent & 50,000 generated images                             & Fixed ADM evaluator references                 \\
        ImageNet probe      & $4\times32\times32$ latent                                                & 12,800 training; 3,200 calibration; 16,000 held out & Disjoint ImageNet training images              \\
        CIFAR-10 generation & Centered $3\times32\times32$ RGB pixels                                   & 50,000 generated images                             & Official 1-Rectified Flow reference statistics \\
        CIFAR-10 probe      & Centered $3\times32\times32$ RGB pixels                                   & 16,000 training; 3,200 calibration; 4,000 held out  & CIFAR-10 training images                       \\
        \bottomrule
    \end{tabularx}
\end{table}
\FloatBarrier{}

\subsection{Samplers}

Sampler definitions do not depend on model architecture.
Table~\ref{tab:sampler-summary} lists each sampler, its update rule, its model-call budget, and where we use it.
\begin{table}[ht]
    \centering
    \small
    \setlength{\tabcolsep}{4pt}
    \caption{
        Sampler variants used in the experiments.
        The budget column gives fixed model-call counts; adaptive RK45 chooses its own number of calls.
    }%
    \label{tab:sampler-summary}
    \begin{tabularx}{\textwidth}{@{}L{2.8cm}>{\raggedright\arraybackslash}X L{1.8cm}L{3.5cm}@{}}
        \toprule
        Sampler                   & Update                                                                                                                           & Budget          & Experiments                                                                    \\
        \midrule
        Hybrid SDE                & SiT SDE\@; $N-1$ stochastic Euler updates through $t=0.96$, then one Mean update; the public setting uses $N=50$~\citep{ma2024sit} & 25--250 calls   & Changed-state response and direction controls; model-time correction; guidance \\
        Uniform stochastic Euler  & Equation~\ref{eq:euler-maruyama} on a uniform grid~\citep{ma2024sit}                                                             & 25--250 calls   & Population time lag and gain grids; changed-state response across model sizes  \\
        Deterministic Euler ODE   & Equation~\ref{eq:euler-ode}                                                                                                      & 50 or 250 calls & ODE response; U-Net and CIFAR-10 transfer                                      \\
        Euler--Maruyama with Mean & LFM SDE with $w_s=\sigma_s$ and one Mean update~\citep{dao2023lfm}                                                               & 25--250 calls   & ADM U-Net transfer                                                             \\
        Velocity-derived SDE      & Equation~\ref{eq:velocity-derived-sde}                                                                                           & 25--250 calls   & CIFAR-10 SDE gain curves and lag calibration                                   \\
        Stochastic Heun           & Predictor--corrector update defined below                                                                                        & 50 calls        & Euler--Heun comparison                                                         \\
        Adaptive RK45             & Official adaptive ODE solver~\citep{liu2022flow}                                                                                 & Adaptive        & CIFAR-10 Euler--RK45 comparison                                                \\
        \bottomrule
    \end{tabularx}
\end{table}
\FloatBarrier{}

\paragraph{Euler updates.}
For the ODE $dX_t=\gamma(t)v_\theta(X_t,t)dt$, deterministic Euler uses
\begin{equation}
    X_{k+1}=X_k+\gamma(t_k)v_\theta(X_k,t_k)\Delta t.
    \label{eq:euler-ode}
\end{equation}
For an SDE $dX_t=b_\gamma(X_t,t)dt+g(t)dW_t$, stochastic Euler, also called Euler--Maruyama, uses
\begin{equation}
    X_{k+1}=X_k+b_\gamma(X_k,t_k)\Delta t
    +g(t_k)\sqrt{\Delta t}\,\xi_k,
    \qquad \xi_k\sim\mathcal N(0,I).
    \label{eq:euler-maruyama}
\end{equation}
The public NFE-50 hybrid sampler uses 49 stochastic Euler updates and one final Mean update~\citep{ma2024sit}.
The Mean update uses the drift but omits the Brownian noise.
The uniform stochastic Euler sampler uses Equation~\ref{eq:euler-maruyama} at evenly spaced times.
The ADM U-Net sampler follows the LFM Euler--Maruyama rule and ends with a noise-free Mean update~\citep{dao2023lfm}.

\paragraph{Stochastic Heun and adaptive RK45.}
Stochastic Heun first takes an Euler--Maruyama predictor.
It then averages the drift and diffusion at the two step endpoints.
The predictor and corrector use the same Brownian increment.
Adaptive RK45 follows the official 1-Rectified Flow implementation with relative and absolute tolerances of $10^{-5}$~\citep{liu2022flow}.

\paragraph{Velocity-derived SDE.}
The CIFAR-10 model predicts velocity rather than score.
We derive the score from that velocity before we form the SDE drift.
For this sampler, let $X_t=tX_1+(1-t)X_0$ with $X_0\sim\mathcal N(0,I)$.
The population velocity and score obey
\begin{equation}
    \nabla_x\log p_t(x)
    =\frac{t v(x,t)-x}{1-t}.
    \label{eq:velocity-derived-score}
\end{equation}
With diffusion $g(t)=1-t$, the drift $v+\tfrac12 {g(t)}^2\nabla_x\log p_t$ has the same Fokker--Planck equation as the probability-flow drift $v$.
Applying gain to the model velocity in this drift gives
\begin{equation}
    dX_t
    =\left[\gamma v_\theta(X_t,t)
        +\frac{1-t}{2}\bigl(t\gamma v_\theta(X_t,t)-X_t\bigr)\right]dt
    +(1-t)dW_t.
    \label{eq:velocity-derived-sde}
\end{equation}
At gain one, insert the population velocity into this SDE\@.
The resulting process follows the prescribed linear-path marginals.
Gain scales only the model velocity terms.
The state term and Brownian noise remain unchanged.

\subsection{Metrics}

\paragraph{FID.}
FID compares the feature mean and covariance of generated and real images; lower values are better~\citep{heusel2017fid}.
The ImageNet experiments use ADM Inception features and the precomputed reference moments supplied with the ADM evaluator~\citep{dhariwal2021adm,ma2024sit}.
The CIFAR-10 experiments use the TF-Hub Inception network and the reference statistics released with 1-Rectified Flow~\citep{liu2022flow}.

\paragraph{sFID.}
Spatial FID compares spatial Inception features; lower values are better~\citep{nash2021sfid}.

\paragraph{Inception Score.}
Inception Score uses predicted class probabilities to measure image confidence and class variety; higher values are better~\citep{salimans2016inceptionscore}.

\paragraph{Precision and recall.}
Precision measures how often generated features lie near real features.
Recall measures how much of the real feature distribution the generated features cover.
Higher values are better for both metrics~\citep{kynkaanniemi2019precision}.
The ADM evaluator supplies 10,000 stored real feature rows for these metrics.
We do not use these rows for FID\@.

\paragraph{Density and coverage.}
Density measures generated concentration in real-feature neighborhoods.
Coverage measures how many real-feature neighborhoods contain a generated sample.
Higher values are better for both metrics, but density must be read with coverage~\citep{naeem2020density}.

\paragraph{DINOv2 Fr\'echet distance.}
This check computes a Fr\'echet distance in ViT-L/14 feature space; lower values are better~\citep{oquab2024dinov2}.

\subsection{Uncertainty}

\paragraph{Pairing.}
A paired comparison uses the same initial state in both settings.
When the model is class conditional, both settings also use the same label.
For a stochastic sampler, both settings also use the same Brownian increments.
The stored image or trajectory rows retain this pairing.

\paragraph{Paired image intervals.}
Each resample selects the same generated-image rows from the two settings.
Paired FID intervals use 2,000 first-order influence resamples with the real reference fixed.

\paragraph{Paired trajectory intervals.}
Trajectory intervals use 2,000 whole-trajectory bootstrap resamples.
Each resample selects the same trajectories from the two settings.

\paragraph{Scope of the intervals.}
The intervals measure generated-sample variation while holding the real reference fixed.

\paragraph{Independent-population check.}
Independent image populations lack the noise reduction from pairing.
We therefore checked the scale of raw FID variation across two independent 50,000-image populations from the same NFE-25 SiT setting.
Both populations use constant gain 1.07216.
Their FIDs are 12.5 and 12.7.
The 0.2 difference shows the scale of between-population FID variation at this sample size.

\Needspace{10\baselineskip}
\section{Population time lag and model-time correction}%
\label{app:state-time-probes}

This section shows that sampled states lag behind genuine states at the same model time.
Velocity scaling reduces this population time lag, and correcting the lag improves image quality.
Velocity scaling gives a larger FID reduction than the tested model-time correction.

\subsection{Probe architecture and training}

\citet{kadkhodaie2026blind} train a noise-level estimator to measure schedule mismatch in diffusion sampling.
We similarly train a probe on genuine path states to estimate path time:
\[
    x_t=(1-t)x_0+tx_1,
    \qquad
    (x_t,y)\longmapsto\widehat s_\phi(x_t,y)\approx t.
\]
The input is a genuine path state $x_t$ and its class label $y$; the output is estimated path time.
For unconditional CIFAR-10, we omit $y$.

\paragraph{Population variation.}
At a fixed path time, genuine states vary with the data example, source noise, and class label when present.
This variation matters because a single probe reading mixes time with ordinary differences between states.
We therefore compare average readings across many paths rather than correcting samples one by one.

\paragraph{Probe architecture.}
Each probe uses three $3\times3$ stride-2 convolutions with 16, 32, and 64 channels and SiLU activations, followed by global average pooling and a linear head.
The head predicts 20 equal-width time bins on $[0,1]$; its posterior mean over the bin centers gives a time estimate in $[0.025,0.975]$.
ImageNet probes standardize latent channels and concatenate the pooled 64-dimensional feature with a 16-dimensional class embedding before the head.
The scale-normalized SiT probe divides each state by its global RMS before channel standardization.
The ADM U-Net probe uses the model's latent space and sampling-time convention.
The CIFAR-10 probe uses standardized centered RGB pixels without a class embedding.

\paragraph{Probe training.}
We train each probe only on genuine states, using batch size 256 and AdamW~\citep{loshchilov2019adamw} with learning rate $10^{-3}$ and weight decay $10^{-4}$.
We use the final fixed epoch without validation selection.
After training, a separate set of genuine states selects one scalar softmax temperature by minimizing cross-entropy on a 401-point log-spaced grid from 0.25 to 4.0.
For the ImageNet SiT analysis, training and saved-state evaluation of the probes took about 30 seconds on 16 CPU threads once the states had been prepared.

\begin{table}[ht]
    \centering
    \small
    \setlength{\tabcolsep}{4pt}
    \caption{
        Probe datasets and held-out accuracy.
        All states come from genuine linear paths; generated states are never used for training or calibration.
        Counts give separate training, probability-calibration, and held-out sets.
        Calibration states are ordinary genuine states used only to select the scalar softmax temperature.
        Held-out error is the mean absolute difference between posterior-mean probe time and known path time on the unit time scale; lower is better.
    }%
    \label{tab:probe-training-summary}
    \begin{tabular*}{\textwidth}{@{\extracolsep{\fill}}lrrrrr@{}}
        \toprule
        Probe & \shortstack{Training\\states} & \shortstack{Calibration\\states} & \shortstack{Held-out\\states} & Epochs & \shortstack{Held-out\\error} \\
        \midrule
        SiT, ImageNet & 12,800 & 3,200 & 16,000 & 30 & 0.0258 \\
        ADM U-Net, ImageNet & 12,800 & 3,200 & 16,000 & 30 & 0.0250 \\
        Rectified Flow, CIFAR-10 & 16,000 & 3,200 & 4,000 & 20 & 0.0141 \\
        \bottomrule
    \end{tabular*}
\end{table}
\FloatBarrier{}

\newpage
\subsection{Defining and validating population time lag}
\label{app:lag-validation}

We next define population time lag and test whether the probe measures progress along the genuine path rather than arbitrary off-path motion.

For a probe $\widehat s_\phi(x,y)$, the genuine-state baseline and estimated population time lag are
\begin{align}
    b_\phi(t)
     & =\mathbb E_{(x,y)\sim p_t}[\widehat s_\phi(x,y)],             \\
    \widehat\delta_\phi(t)
     & =b_\phi(t)-\mathbb E_{(x,y)\sim q_t^u}[\widehat s_\phi(x,y)].
\end{align}
Because the probe estimates time from 20 bins with temperature-scaled probabilities, its average reading can have a systematic, time-dependent bias.
We therefore measure lag by comparing sampled states with genuine states at the same time, so the probe's bias is not counted as sampling lag.
On 16,000 held-out genuine states per model, the mean absolute deviation of individual readings from their genuine-state population curve is 0.0226 for the SiT raw-state probe and 0.0216 for the ADM raw-state probe.
For the scale-normalized SiT probe, it is 0.0415 on 3,200 held-out genuine states at the four Euler--Heun saved times.
These measurements support the population averaging described above.
\Needspace{6\baselineskip}
\paragraph{Path-direction control.}
We tested whether the SiT raw-state probe detects a known reduction in the time $t$ used to construct a genuine state.
For each of 11,200 held-out ImageNet states $X_t$, we kept the noise sample $X_0$ and data sample $X_1$ fixed.
Reducing $t$ by $\Delta t$ gives
\[
    X_{t-\Delta t}=(1-t+\Delta t)X_0+(t-\Delta t)X_1.
\]

For comparison, we drew a standard Gaussian vector for each state.
We projected the vector perpendicular to the path direction $U=X_1-X_0$.
We rescaled it to match the $L_2$ length of $X_{t-\Delta t}-X_t$.
We then added this vector to the original state $X_t$.

\begin{table}[ht]
    \centering
    \small
    \setlength{\tabcolsep}{4pt}
    \caption{
        SiT raw-state probe responses on 11,200 held-out genuine ImageNet states.
        Entries give mean changes in estimated time on the unit time scale, measured as the perturbed estimate minus the original estimate.
        The difference is the perpendicular response minus the response to reducing $t$.
        Intervals use 2,000 paired bootstrap resamples within time bins.
        A response close to $-\Delta t$ after reducing $t$, together with a perpendicular response near zero, supports sensitivity to path time.
    }%
    \label{tab:probe-path-direction}
    \begin{tabular}{@{}rrrr@{}}
        \toprule
        \shortstack{Reduction\\in $t$} & \shortstack{Probe change from\\reducing $t$} & \shortstack{Probe change from\\perpendicular perturbation} & \shortstack{Difference\\{}[95\% interval]} \\
        \midrule
        0.025 & $-0.0248$ & $-0.00130$ & $+0.0235\;[+0.0234,+0.0236]$ \\
        0.050 & $-0.0499$ & $-0.00519$ & $+0.0447\;[+0.0445,+0.0449]$ \\
        \bottomrule
    \end{tabular}
\end{table}
\FloatBarrier{}

The probe tracks the known changes in time and responds much less to equally large perpendicular perturbations (Table~\ref{tab:probe-path-direction}).

\subsection{Coarse sampling and missing refinement}%
\label{app:coarse-refinement}

Assume independent noise and data conditional on class $y$, and let $\mu_y=\mathbb E[X_1\mid y]$.
The initial population field is
\begin{equation}
    v^*(x,0,y)
    =\mathbb E[X_1-X_0\mid X_0=x,y]
    =\mu_y-x.
\end{equation}
One Euler step from noise to data therefore gives
\begin{equation}
    \widetilde X_1=X_0+v^*(X_0,0,y)=\mu_y.
\end{equation}
Every initial state reaches the same class mean, with no sample-specific variation.

With two Euler steps (NFE 2), the state after the first step is
\begin{equation}
    \widetilde X_{1/2}=\tfrac12X_0+\tfrac12\mu_y,
\end{equation}
whereas a genuine training-path state at the same time is
\begin{equation}
    X_{1/2}=\tfrac12X_0+\tfrac12X_1.
\end{equation}
The Euler state has the expected source noise amplitude but lacks the sample-specific data structure present in genuine states.
The comparison used 11,200 paired ImageNet latent states at $t=0.5$, with the same source noise and class label in each pair.
We estimated each class mean from 16 separate images.
To focus on image structure, we used a probe trained and evaluated on states divided by their own global RMS.

The mean estimated time was 0.254 for the Euler states and 0.493 for genuine states.
The Euler-minus-genuine difference was $-0.240$, with a paired 95\% bootstrap interval of $[-0.241,-0.238]$.
The interval uses 2,000 paired resamples and conditions on the trained probe and estimated class means.

\subsection{Model-time correction and scale control}
\label{app:model-time-correction}

\paragraph{Model-time correction improves image quality.}
At each model call, we replaced $t$ with $t-\widehat\delta_\phi(t)$ using the population time lag measured by the raw-state probe.
We left the integration clock and velocity scale unchanged.
At NFE 50, this correction reduced FID from 13.8 to 8.8 under the unguided public hybrid SDE sampler.
Both settings used 50,000 images with the same class labels, initial latents, and Brownian paths.

\paragraph{A separate probe tests the role of global scale.}
We trained this probe on states divided by their own global RMS and used the same normalization at evaluation.
It otherwise uses the same architecture and genuine-state training examples as the raw-state probe.
The population time lag remained.
On 1,024 unscaled paths from uniform stochastic Euler sampling at NFE 25, the scale-normalized population time lag peaked at 0.0453 and ended at 0.0152 on the unit time scale.

\FloatBarrier{}

At NFE 50, the scale-normalized time correction reduced FID from 13.8 to 8.6 on 50,000 exactly paired images.
The correction-minus-nominal FID contrast was $-5.2\;[-5.3,-5.0]$ with a paired 95\% interval.
The raw-state correction reached 8.8.
The similar results show that the timing effect is not just a norm readout.

\Needspace{8\baselineskip}
\subsection{Model-time correction transfers to an ADM U-Net}
\label{app:unet-model-time-correction}
We test whether correcting population time lag also improves image quality in a different model architecture.

At NFE 50, a fixed correction derived from the U-Net's raw-state population time lag changed only the time passed to the network.
The intervention kept unit velocity gain and every sampler term nominal.

\begin{table}[ht]
    \centering
    \small
    \setlength{\tabcolsep}{5pt}
    \caption{
    ADM U-Net image quality under the LFM Euler--Maruyama sampler described in Appendix~\ref{app:design} at NFE 50.
    Lower FID is better.
    Each method contains 50,000 ImageNet-256 images with exact class-label, initial-latent, and Brownian pairing.
    Relative to unscaled sampling, model-time correction changed FID by $-6.0\;[-6.2,-5.9]$; relative to gain 1.075, it changed FID by $+3.5\;[+3.4,+3.7]$.
    Each contrast is model-time correction minus the other method, and brackets are paired 95\% percentile intervals from 2,000 common-row influence resamples with the real reference fixed.
    }%
    \label{tab:unet-time-correction}
    \begin{tabular}{@{}lr@{}}
        \toprule
        Method                       & FID  \\
        \midrule
        Unscaled sampling            & 22.7 \\
        Model-time correction        & 16.7 \\
        Velocity scaling, gain 1.075 & 13.2 \\
        \bottomrule
    \end{tabular}
\end{table}
\FloatBarrier{}

Model-time correction improves ADM U-Net FID, but velocity scaling at gain 1.075 improves it further.

\Needspace{12\baselineskip}
\section{Scalar lag calibration and image quality}%
\label{app:scalar-calibration}

\subsection{Direct scalar lag calibration}
\label{app:scalar-lag-calibration}

Let $\widehat\delta_\phi(t;\gamma)$ denote Equation~\ref{eq:lag} evaluated on states sampled with constant gain $\gamma$.
For the $K$ post-update times $t_i$, we select the lag-based gain by
\begin{equation}
    \gamma_{\rm lag}
    =\arg\min_{\gamma>0}\frac{1}{K}\sum_{i=1}^{K}
    \widehat\delta_\phi(t_i;\gamma)^2.
    \label{eq:probe-gain}
\end{equation}
Each gain uses the same 1,280 trajectories within a model and NFE, with fixed initial states, class labels, and Brownian paths.
All four SiT sizes use uniform stochastic Euler sampling.
The ADM U-Net uses Euler--Maruyama through time 0.96 and a fixed final Mean step of length 0.04.
The CIFAR-10 model uses the corrected pixel-space Euler--Maruyama sampler in Equation~\ref{eq:velocity-derived-sde}.
The objective averages all post-update times equally, including the endpoint.

We bracket the lag minimum with three evaluated gains whose middle gain gives the lowest objective value.
We use SciPy's implementation of Brent's scalar minimization method~\citep{brent1973minimization}, which combines parabolic interpolation with golden-section search.
We evaluate each proposed gain on the same paired trajectories.
We select the evaluated gain with the lowest objective value.

The SiT-XL/2 exploration included a grid from 0.975 to 1.175 in steps of 0.0125.
Existing schedule averages and FID-selected gains supplied additional candidate values where available; their FID values never entered the lag objective.
A separate calibration check used discrete slope bisection at gain resolution 0.01, starting from gains 1.00, 1.04, and 1.08 (Appendix~\ref{app:calibration-cost}).

Brent's relative gain tolerance was 0.001, except 0.0001 for SiT-XL/2 at NFE 25.
For ADM at NFE 250, the observed probe response motivated additional paired candidates at 1.007, 1.008, and 1.009, evaluated independently before the final refinement.
The finer numerical tolerance does not measure statistical precision.
We use 2,000 paired whole-trajectory bootstrap draws to estimate uncertainty, selecting the best gain among the evaluated candidates in each draw.

The gain intervals condition on the trained probe and the genuine-state baseline.
They do not include uncertainty from retraining the probe or from optimizing between measured candidates.
Table~\ref{tab:scalar-lag-calibration} reports lag on the gain-selection paths.
Table~\ref{tab:scalar-lag-validation} evaluates each fixed lag-based gain on 1,280 separate paired trajectories.

\begin{table}[ht]
\centering
\small
\setlength{\tabcolsep}{4pt}
\caption{
Direct scalar lag calibration across the six models at NFE 25, 50, 100, and 250.
SiT and ADM use unguided ImageNet-256 sampling; NCSN++ uses CIFAR-10.
Gains are dimensionless; brackets give paired 95\% gain intervals from 2,000 whole-trajectory draws over 1,280 paths per candidate, with the trained probe and genuine baseline fixed.
The lag columns give RMS population probe-time mismatch multiplied by $10^3$; lower is better.
Unscaled uses gain one, scalar uses the lag-based gain, and lowest-FID uses the lowest measured FID gain from a 50,000-image grid.
These paths were used for gain selection; Table~\ref{tab:scalar-lag-validation} gives separate fresh-path validation.
SiT uses uniform stochastic Euler; ADM uses Euler--Maruyama with a final Mean step; CIFAR-10 uses Equation~\ref{eq:velocity-derived-sde}.
}
\label{tab:scalar-lag-calibration}
\begin{tabular}{@{}lrccrrr@{}}
\toprule
Model & NFE & \shortstack{Lag-based gain\\{[95\%]}} & \shortstack{Lowest-FID\\gain} & \shortstack{Unscaled\\lag RMS} & \shortstack{Scalar\\lag RMS} & \shortstack{Lowest-FID\\lag RMS} \\
\midrule
SiT-S/2 & 25 & $1.074\;[1.070,1.077]$ & 1.150 & 25.8 & 6.0 & 23.9 \\
 & 50 & $1.033\;[1.029,1.034]$ & 1.100 & 12.1 & 3.7 & 21.9 \\
 & 100 & $1.014\;[1.010,1.017]$ & 1.075 & 5.4 & 2.4 & 20.4 \\
 & 250 & $1.004\;[1.000,1.007]$ & 1.050 & 2.7 & 2.3 & 16.1 \\
SiT-B/2 & 25 & $1.073\;[1.070,1.077]$ & 1.125 & 24.9 & 5.4 & 16.5 \\
 & 50 & $1.032\;[1.029,1.034]$ & 1.075 & 11.5 & 3.3 & 14.0 \\
 & 100 & $1.013\;[1.010,1.017]$ & 1.050 & 5.2 & 2.3 & 12.4 \\
 & 250 & $1.004\;[1.000,1.007]$ & 1.050 & 2.6 & 2.2 & 15.7 \\
SiT-L/2 & 25 & $1.073\;[1.069,1.077]$ & 1.125 & 24.2 & 4.8 & 16.2 \\
 & 50 & $1.031\;[1.029,1.034]$ & 1.075 & 10.8 & 2.7 & 14.0 \\
 & 100 & $1.012\;[1.010,1.017]$ & 1.050 & 4.6 & 1.9 & 12.4 \\
 & 250 & $1.002\;[1.000,1.007]$ & 1.050 & 2.3 & 2.1 & 15.7 \\
SiT-XL/2 & 25 & $1.077\;[1.073,1.080]$ & 1.150 & 25.4 & 5.8 & 21.6 \\
 & 50 & $1.034\;[1.033,1.038]$ & 1.100 & 12.1 & 3.9 & 19.8 \\
 & 100 & $1.017\;[1.012,1.017]$ & 1.075 & 6.1 & 2.6 & 18.0 \\
 & 250 & $1.007\;[1.005,1.010]$ & 1.050 & 3.7 & 2.9 & 13.9 \\
ADM U-Net & 25 & $1.077\;[1.075,1.079]$ & 1.125 & 25.3 & 6.3 & 15.2 \\
 & 50 & $1.036\;[1.031,1.037]$ & 1.075 & 12.5 & 4.4 & 12.8 \\
 & 100 & $1.016\;[1.014,1.017]$ & 1.075 & 6.6 & 3.6 & 18.7 \\
 & 250 & $1.008\;[1.007,1.009]$ & 1.050 & 4.0 & 3.0 & 13.9 \\
CIFAR NCSN++ & 25 & $1.018\;[1.017,1.019]$ & 1.000 & 6.0 & 3.1 & 6.0 \\
 & 50 & $1.007\;[1.006,1.008]$ & 1.000 & 3.0 & 2.2 & 3.0 \\
 & 100 & $1.002\;[1.000,1.004]$ & 1.000 & 1.8 & 1.8 & 1.8 \\
 & 250 & $0.999\;[0.998,1.000]$ & 1.000 & 1.5 & 1.5 & 1.5 \\
\bottomrule
\end{tabular}
\end{table}

\FloatBarrier{}

Appendix~\ref{app:gain-fid-models} reports FID at these gains and compares gain-prediction rules.

\Needspace{8\baselineskip}
\begin{minipage}{\textwidth}
\subsection{Scalar validation across models and NFE values}
The fixed lag-based gains reduce measured population time lag on fresh paths in all twenty ImageNet settings.
At NFE 250, the reductions for SiT-S/2, B/2, and L/2 are small and their uncertainty includes zero (Table~\ref{tab:scalar-lag-validation}).

CIFAR-10 has little population time lag.
Its selected gains reduce lag on fresh paths at NFE 25, 50, and 100; at NFE 250, lag rises slightly, with uncertainty including zero.
The lowest-FID gain remains one at all four NFEs.
At NFE 50, the interpolated FID at the lag-based gain is 4.1, versus 3.0 for unscaled sampling (Table~\ref{tab:lowest-fid-grid-summary}).

\captionsetup{type=table}
\centering
\small
\setlength{\tabcolsep}{4pt}
\captionof{table}{
Fresh-path validation of the fixed lag-based gains in Table~\ref{tab:scalar-lag-calibration}.
Each setting uses 1,280 new paired trajectories per arm, with the gain, trained probe, and genuine baseline fixed from calibration.
Values give RMS population probe-time mismatch multiplied by $10^3$; lower is better.
The reduction is unscaled minus scalar lag; positive values favor the scalar.
Brackets are paired 95\% whole-trajectory bootstrap intervals from 2,000 draws.
Models and samplers match Table~\ref{tab:scalar-lag-calibration}; no image metric is used.
}
\label{tab:scalar-lag-validation}
\begin{tabular}{@{}lrrrr@{}}
\toprule
Model & NFE & Unscaled lag RMS & Scalar lag RMS & Reduction [95\%] \\
\midrule
SiT-S/2 & 25 & 26.4 & 6.5 & $+19.9\;[+19.0,+20.8]$ \\
 & 50 & 12.7 & 4.2 & $+8.5\;[+7.6,+9.2]$ \\
 & 100 & 6.2 & 2.7 & $+3.5\;[+2.7,+3.9]$ \\
 & 250 & 2.8 & 2.5 & $+0.3\;[-0.3,+0.7]$ \\
SiT-B/2 & 25 & 25.9 & 6.0 & $+19.9\;[+19.0,+20.7]$ \\
 & 50 & 12.4 & 4.1 & $+8.3\;[+7.5,+9.0]$ \\
 & 100 & 6.0 & 2.7 & $+3.3\;[+2.6,+3.7]$ \\
 & 250 & 2.7 & 2.5 & $+0.2\;[-0.3,+0.7]$ \\
SiT-L/2 & 25 & 25.4 & 5.8 & $+19.6\;[+18.7,+20.4]$ \\
 & 50 & 12.0 & 3.8 & $+8.2\;[+7.4,+8.8]$ \\
 & 100 & 5.6 & 2.5 & $+3.1\;[+2.4,+3.5]$ \\
 & 250 & 2.8 & 2.6 & $+0.3\;[-0.1,+0.5]$ \\
SiT-XL/2 & 25 & 25.8 & 6.3 & $+19.6\;[+18.7,+20.4]$ \\
 & 50 & 13.0 & 4.9 & $+8.1\;[+7.3,+8.7]$ \\
 & 100 & 7.2 & 3.8 & $+3.4\;[+2.7,+3.9]$ \\
 & 250 & 4.0 & 3.1 & $+0.9\;[+0.3,+1.3]$ \\
ADM U-Net & 25 & 26.0 & 7.0 & $+19.0\;[+18.1,+19.9]$ \\
 & 50 & 12.9 & 5.2 & $+7.7\;[+6.8,+8.5]$ \\
 & 100 & 7.8 & 3.9 & $+3.9\;[+3.3,+4.3]$ \\
 & 250 & 4.4 & 3.2 & $+1.2\;[+0.6,+1.6]$ \\
CIFAR NCSN++ & 25 & 6.5 & 3.1 & $+3.4\;[+3.1,+3.7]$ \\
 & 50 & 3.4 & 2.2 & $+1.2\;[+0.9,+1.4]$ \\
 & 100 & 2.1 & 1.8 & $+0.2\;[+0.1,+0.3]$ \\
 & 250 & 1.5 & 1.6 & $-0.1\;[-0.2,+0.0]$ \\
\bottomrule
\end{tabular}
\end{minipage}

\FloatBarrier{}

\Needspace{8\baselineskip}
\begin{minipage}{\textwidth}
\subsection{Gain and FID across models}
\label{app:gain-fid-models}
Lag-based gains recover much of the ImageNet FID improvement at low NFE, while FID favors additional gain at every tested NFE.
Table~\ref{tab:lowest-fid-grid-summary} compares all six models and four NFEs, using interpolated lag-gain FIDs and measured FIDs at the closest tested $\kappa$-adjusted gains.
For CIFAR-10, gain one gives the lowest measured FID throughout.

\captionsetup{type=table}
\centering
\small
\setlength{\tabcolsep}{4pt}
\captionof{table}{
FID and selected gains across all six models and four NFEs.
The lag-based column gives the fitted gain and its linearly interpolated FID; interpolation uses the unrounded gain and its two bracketing measured gains.
The $\kappa$-adjusted column gives the closest tested gain to the full-gain prediction and its measured FID; $\kappa$ is fitted at NFE 25 and reused at higher NFEs.
No additional multiplier is fitted for CIFAR-10 (---), where gain one has the lowest measured FID.
Every measured gain uses 50,000 paired images within its model and NFE, and lower FID is better.
SiT uses uniform stochastic Euler, ADM uses Euler--Maruyama with a final Mean step, and CIFAR-10 uses Equation~\ref{eq:velocity-derived-sde}.
ImageNet and CIFAR-10 use different evaluators and reference distributions; their FIDs are not comparable.
These lookup and interpolation summaries have no computed uncertainty intervals.
}
\label{tab:lowest-fid-grid-summary}
\begin{tabular}{@{}lrrrrr@{}}
\toprule
Model & NFE & \shortstack{Unscaled\\FID} & \shortstack{Lag-based gain,\\interpolated FID} & \shortstack{$\kappa$-adjusted\\tested gain, FID} & \shortstack{Lowest-FID\\gain, FID} \\
\midrule
SiT-S/2 & 25 & 71.1 & 1.074, 50.9 & 1.150, 44.4 & 1.150, 44.4 \\
 & 50 & 58.3 & 1.033, 50.2 & 1.100, 43.7 & 1.100, 43.7 \\
 & 100 & 53.3 & 1.014, 50.3 & 1.075, 43.8 & 1.075, 43.8 \\
 & 250 & 50.6 & 1.004, 49.9 & 1.075, 44.1 & 1.050, 43.8 \\
SiT-B/2 & 25 & 47.9 & 1.073, 30.2 & 1.125, 25.9 & 1.125, 25.9 \\
 & 50 & 35.8 & 1.032, 29.2 & 1.075, 25.2 & 1.075, 25.2 \\
 & 100 & 31.5 & 1.013, 29.3 & 1.050, 25.2 & 1.050, 25.2 \\
 & 250 & 29.1 & 1.004, 28.5 & 1.050, 24.8 & 1.050, 24.8 \\
SiT-L/2 & 25 & 30.2 & 1.073, 16.1 & 1.125, 13.4 & 1.125, 13.4 \\
 & 50 & 19.6 & 1.031, 14.9 & 1.075, 12.5 & 1.075, 12.5 \\
 & 100 & 16.1 & 1.012, 14.7 & 1.050, 12.3 & 1.050, 12.3 \\
 & 250 & 14.6 & 1.002, 14.4 & 1.050, 12.2 & 1.050, 12.2 \\
SiT-XL/2 & 25 & 28.0 & 1.077, 12.2 & 1.150, 8.7 & 1.150, 8.7 \\
 & 50 & 15.8 & 1.034, 10.2 & 1.100, 6.9 & 1.100, 6.9 \\
 & 100 & 11.5 & 1.017, 9.5 & 1.075, 6.6 & 1.075, 6.6 \\
 & 250 & 9.7 & 1.007, 9.1 & 1.075, 6.6 & 1.050, 6.3 \\
ADM U-Net & 25 & 34.8 & 1.077, 17.9 & 1.125, 14.7 & 1.125, 14.7 \\
 & 50 & 22.7 & 1.036, 16.4 & 1.075, 13.2 & 1.075, 13.2 \\
 & 100 & 18.4 & 1.016, 16.0 & 1.050, 13.1 & 1.075, 12.9 \\
 & 250 & 16.8 & 1.008, 15.8 & 1.050, 12.8 & 1.050, 12.8 \\
CIFAR NCSN++ & 25 & 3.5 & 1.018, 4.5 & --- & 1.000, 3.5 \\
 & 50 & 3.0 & 1.007, 4.1 & --- & 1.000, 3.0 \\
 & 100 & 3.0 & 1.002, 3.2 & --- & 1.000, 3.0 \\
 & 250 & 2.9 & 0.999, 3.0 & --- & 1.000, 2.9 \\
\bottomrule
\end{tabular}
\end{minipage}
\FloatBarrier{}

\Needspace{8\baselineskip}
\begin{minipage}{\textwidth}
    \paragraph{ADM U-Net gain curves.}

    Figure~\ref{fig:unet-sde-gain} shows the full ADM U-Net gain curves underlying Table~\ref{tab:lowest-fid-grid-summary}.
    At every NFE, paired 95\% intervals show that the lowest-FID gain performs better than both adjacent tested gains; Table~\ref{tab:unet-sde-gain} gives the contrasts with the next higher gain.

    \centering
    \includegraphics[width=\textwidth]{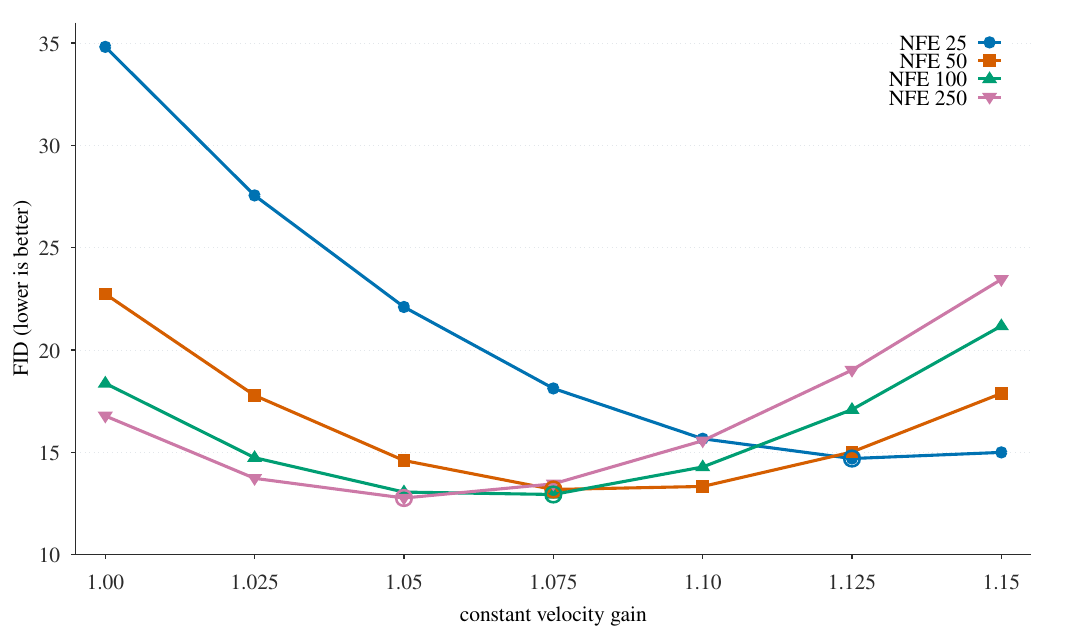}
    \captionsetup{type=figure}
    \captionof{figure}{
        FID response to constant gain for the released ImageNet-256 LFM ADM U-Net under the Euler--Maruyama sampler described in Appendix~\ref{app:design}.
        Lower is better; open markers indicate the lowest measured FID on each seven-point grid.
        Each point uses 50,000 images, with exact class-label, initial-latent, and Brownian pairing within each NFE\@.
        The minimum moves toward gain one as NFE rises.
    }%
    \label{fig:unet-sde-gain}
\end{minipage}

\Needspace{22\baselineskip}
\paragraph{Neighboring gains.}
Table~\ref{tab:unet-sde-gain} gives the FID increase at the tested gain immediately above each minimum in Figure~\ref{fig:unet-sde-gain}.

\begin{table}[ht]
    \centering
    \small
    \setlength{\tabcolsep}{6pt}
    \caption{
        FID at the lowest-FID gain and next higher tested gain for the unguided ImageNet-256 ADM U-Net under LFM Euler--Maruyama sampling with a deterministic final Mean step; lower FID is better.
        The lowest-FID gain minimizes measured FID over the grid in Figure~\ref{fig:unet-sde-gain}.
        The contrast is $\Delta\mathrm{FID}=\mathrm{FID}_{\rm higher}-\mathrm{FID}_{\rm lowest}$, so positive values favor the lowest-FID gain.
        Contrasts use unrounded FIDs; brackets give paired 95\% intervals from 2,000 common-row influence resamples with the real reference fixed.
        Each gain uses 50,000 images paired by class label, initial latent, and Brownian path within each NFE\@.
    }%
    \label{tab:unet-sde-gain}
    \begin{tabular}{@{}rrrrrr@{}}
        \toprule
        NFE & \shortstack{Lowest-FID\\gain} & FID & \shortstack{Next higher\\gain} & FID & $\Delta$FID [95\%] \\
        \midrule
        25  & 1.125 & 14.7 & 1.150 & 15.0 & $+0.3\;[+0.2,+0.4]$ \\
        50  & 1.075 & 13.2 & 1.100 & 13.3 & $+0.1\;[+0.1,+0.3]$ \\
        100 & 1.075 & 12.9 & 1.100 & 14.3 & $+1.3\;[+1.2,+1.4]$ \\
        250 & 1.050 & 12.8 & 1.075 & 13.5 & $+0.7\;[+0.6,+0.8]$ \\
        \bottomrule
    \end{tabular}
\end{table}

\FloatBarrier{}

\Needspace{24\baselineskip}
\paragraph{FID recovery at the lag-based gain.}
The lag-based gains recover an estimated 49\% of the FID reduction from gain one to the lowest-FID gain, averaged equally across the twenty ImageNet settings.
Recovery falls from 81\% at NFE 25 to 15\% at NFE 250 (Table~\ref{tab:scalar-fid-recovery}).

\begin{table}[ht]
\centering
\small
\setlength{\tabcolsep}{9pt}
\caption{
Estimated FID recovery (\%) at the lag-based gains for unguided ImageNet-256 sampling; larger is better.
Recovery is $[\mathrm{FID}(1)-\widehat{\mathrm{FID}}(\gamma_{\rm lag})]/[\mathrm{FID}(1)-\mathrm{FID}(\gamma^\star_{\rm FID})]$.
Recovery is computed from unrounded FIDs.
Table~\ref{tab:lowest-fid-grid-summary} gives the unscaled, interpolated lag-gain, and lowest measured FIDs.
Each measured FID uses 50,000 paired images; SiT uses uniform stochastic Euler and ADM uses Euler--Maruyama with a final Mean step.
The means weight models equally.
No FID or interpolation-error intervals have been computed.
}
\label{tab:scalar-fid-recovery}
\begin{tabular}{@{}lrrrr@{}}
\toprule
Model & NFE 25 & NFE 50 & NFE 100 & NFE 250 \\
\midrule
SiT-S/2 & 76 & 55 & 32 & 11 \\
SiT-B/2 & 81 & 62 & 36 & 13 \\
SiT-L/2 & 84 & 66 & 36 & 9 \\
SiT-XL/2 & 82 & 63 & 42 & 19 \\
ADM U-Net & 84 & 67 & 43 & 24 \\
\midrule
Mean & 81 & 63 & 38 & 15 \\
\bottomrule
\end{tabular}
\end{table}
\FloatBarrier{}

\Needspace{28\baselineskip}
\paragraph{Gain predictions across NFE.}
We fit the multiplier $\kappa$ at NFE 25 and reuse it at higher NFEs, as in Equation~\ref{eq:additional-gain}.
Table~\ref{tab:scalar-gain-predictions} compares the predicted gains with the lowest-FID grid gains.
For SiT-XL/2, $\kappa$ predicts gains with estimated FID within 0.3 of the lowest measured value at NFE 50, 100, and 250.

\begin{table}[ht]
\centering
\small
\setlength{\tabcolsep}{4pt}
\caption{
Full-gain predictions for the five unguided ImageNet models.
Each model's multiplier is fitted at NFE 25 and reused at the three higher NFEs.
The signed error is predicted gain minus lowest-FID grid gain; smaller absolute error is better.
Each FID grid point uses 50,000 paired images within its model and NFE.
These dimensionless gain-error summaries have no computed uncertainty intervals and do not measure FID at the predicted gains.
The SiT-XL/2 FID estimates quoted in the text use linear interpolation at NFE 50 and 100 and the nearest tested gain at NFE 250.
}
\label{tab:scalar-gain-predictions}
\begin{tabular}{@{}lrcrr@{}}
\toprule
Model & NFE & Predicted gain & Lowest-FID gain & Signed error \\
\midrule
SiT-S/2 & 50 & 1.107 & 1.100 & $+0.007$ \\
 & 100 & 1.086 & 1.075 & $+0.011$ \\
 & 250 & 1.076 & 1.050 & $+0.026$ \\
SiT-B/2 & 50 & 1.082 & 1.075 & $+0.007$ \\
 & 100 & 1.062 & 1.050 & $+0.012$ \\
 & 250 & 1.053 & 1.050 & $+0.003$ \\
SiT-L/2 & 50 & 1.082 & 1.075 & $+0.007$ \\
 & 100 & 1.062 & 1.050 & $+0.012$ \\
 & 250 & 1.052 & 1.050 & $+0.002$ \\
SiT-XL/2 & 50 & 1.105 & 1.100 & $+0.005$ \\
 & 100 & 1.086 & 1.075 & $+0.011$ \\
 & 250 & 1.075 & 1.050 & $+0.025$ \\
ADM U-Net & 50 & 1.082 & 1.075 & $+0.007$ \\
 & 100 & 1.061 & 1.075 & $-0.014$ \\
 & 250 & 1.052 & 1.050 & $+0.002$ \\
\bottomrule
\end{tabular}
\end{table}
\FloatBarrier{}

\Needspace{24\baselineskip}
\paragraph{Dependence on the gain rule.}
We compare multiplying the lag-based gain with multiplying its excess above one.
Each rule fits one coefficient at NFE 25 and predicts the three higher NFEs.
The full-gain multiplier has the smallest mean absolute gain error for each ImageNet model (Table~\ref{tab:scalar-gain-rules}).
Across the four SiT sizes, its mean absolute gain error is 0.011 over the twelve higher-NFE predictions.

The factorization describes the observed gains; it does not establish two independent causal mechanisms or a limiting multiplier under finer integration.

\begin{table}[ht]
\centering
\small
\setlength{\tabcolsep}{4pt}
\caption{
Gain-rule comparison for the five unguided ImageNet models.
Each rule fits one coefficient at NFE 25: $\kappa\gamma_{\rm lag}$ or $1+\alpha(\gamma_{\rm lag}-1)$.
The two error columns give the mean of the absolute differences between predicted gain and lowest-FID grid gain at NFE 50, 100, and 250; lower is better.
All gains are dimensionless and use the unrounded scalar fits and lowest-FID gains in Table~\ref{tab:scalar-lag-calibration}.
These summaries have no computed uncertainty intervals.
}
\label{tab:scalar-gain-rules}
\begin{tabular}{@{}lrrr@{}}
\toprule
Model & $\kappa$ & Full-gain error & Excess error \\
\midrule
SiT-S/2 & 1.071 & 0.014 & 0.040 \\
SiT-B/2 & 1.048 & 0.007 & 0.030 \\
SiT-L/2 & 1.049 & 0.007 & 0.032 \\
SiT-XL/2 & 1.068 & 0.014 & 0.037 \\
ADM U-Net & 1.044 & 0.008 & 0.034 \\
\bottomrule
\end{tabular}
\end{table}

\FloatBarrier{}

\Needspace{8\baselineskip}
\subsection{Calibration cost}
\label{app:calibration-cost}

At SiT-XL/2 NFE 50, direct scalar calibration at gain resolution 0.01 used six candidate gains with 256 paired paths each.
This totals 1,536 trajectories and 76,800 model evaluations.
All three probe-seed fits reuse these candidate trajectories.
Separate validation evaluated identity and the selected gain on 1,280 new paired paths per gain; that work is not part of calibration cost.

Replaying 0.01-resolution bisection on the saved lag curves takes 4--9 candidate evaluations across the 24 model/NFE settings, with a median of six.
At 256 paths per candidate, this corresponds to 1,024--2,304 trajectories within the already explored gain brackets.
These replay counts exclude discovering a bracket.
The broader exploration used up to 25 evaluations of 1,280 paths each; that exploration budget is not required by the coarse search.

By comparison, one 50,000-image NFE-50 setting used 2.5 million model evaluations and a median 4.7 RTX-4090 hours.
The full seven-gain sweep used 33 GPU-hours before image-metric extraction.
Scalar calibration requires neither image decoding nor image-feature extraction.

Preparing each 16,000-state genuine calibration and held-out set took about 147 seconds on two RTX 4090 GPUs.
State preparation and probe training are one-time costs reused across model sizes and NFEs.

\Needspace{28\baselineskip}
\subsection{Calibration stability}
\label{app:calibration-stability}

We test how direct scalar gain selection changes with trajectory count and probe-training seed.

\paragraph{Calibration size.}
Using fewer paths changes the fitted scalar only slightly.
Across all 24 settings, 256-path subsets differ from the full 1,280-path gain by 0.00223 on average; 512 paths reduce this difference to 0.00137.
Table~\ref{tab:lag-calibration-convergence} reports the paired subsampling calculation on the saved candidate curves.

\begin{table}[ht]
\centering
\small
\setlength{\tabcolsep}{4pt}
\caption{
Sensitivity of scalar gain selection to trajectory count across all six models and four NFEs.
For each setting and count, 200 paired subsets of the saved 1,280-path candidate bank are used to refit the lag minimum on interpolated population lag curves.
The second column is the mean absolute difference from the 1,280-path fitted gain, multiplied by $10^3$, averaged equally over the 24 settings and 200 subsets; lower means greater stability.
The trained probe and genuine baseline are fixed.
These subsampling summaries have no computed uncertainty intervals and do not include fresh adaptive candidate selection.
}
\label{tab:lag-calibration-convergence}
\begin{tabular}{@{}rr@{}}
\toprule
Paths per candidate & Mean absolute gain change $\times10^3$ \\
\midrule
128 & 3.30 \\
256 & 2.23 \\
512 & 1.37 \\
1024 & 0.55 \\
\bottomrule
\end{tabular}
\end{table}

\paragraph{Probe-seed stability.}
All three SiT-XL/2 probes select gain 1.04 at NFE 50 when the gain resolution is 0.01.
Each selected gain also reduces population time lag on separate fresh paths (Table~\ref{tab:probe-seed-stability}).

\begin{table}[ht]
\centering
\small
\setlength{\tabcolsep}{4pt}
\caption{
Direct scalar calibration with three probe-training seeds for unguided SiT-XL/2 at NFE 50.
Each probe selects a gain at resolution 0.01 from shared candidates evaluated on 256 paired calibration paths.
The identity and selected gains are evaluated on 1,280 separate fresh paired paths; the genuine baseline is recomputed for each probe.
Lag is RMS population probe-time mismatch multiplied by $10^3$, and lower is better.
The reduction is unscaled minus scalar lag; brackets give paired 95\% whole-trajectory bootstrap intervals from 2,000 draws.
Equal selections at 0.01 resolution do not imply zero variation in the continuous optimum.
No image metric is used.
}
\label{tab:probe-seed-stability}
\begin{tabular}{@{}rrrrr@{}}
\toprule
Training seed & Gain & Unscaled lag RMS & Scalar lag RMS & Reduction [95\%] \\
\midrule
2026072307 & 1.04 & 13.6 & 5.0 & $+8.6\;[+7.7,+9.4]$ \\
2026072308 & 1.04 & 12.9 & 4.2 & $+8.7\;[+7.6,+9.6]$ \\
2026072309 & 1.04 & 11.8 & 4.4 & $+7.5\;[+6.4,+8.4]$ \\
\bottomrule
\end{tabular}
\end{table}

\Needspace{8\baselineskip}
\subsection{Schedule shape and scalar calibration}
\label{app:lag-gain-construction}

For unguided SiT-XL/2 under uniform stochastic Euler sampling, a fitted gain schedule and a constant gain both remove most of the population time lag at low NFE (Table~\ref{tab:lag-schedule-average}).

We write the schedule as $\gamma_{\boldsymbol\beta}(t)=1+\sum_{j=1}^6\beta_j B_j(t)$, where $B_j$ are triangular basis functions at six equally spaced times.
For each basis function, paired trajectories at gains $1\pm0.025 B_j$ estimate a column of the response matrix $R$ by central differences.
Then $(R\boldsymbol\beta)_i$ predicts the change in mean probe time at post-update time $t_i$, for $i=1,\ldots,K$.
We fit the coefficients to offset the unscaled lag:
\begin{equation}
    \widehat{\boldsymbol\beta}
    =\arg\min_{\boldsymbol\beta}
    \frac{1}{K}\sum_{i=1}^K
    {\left[\widehat\delta_\phi(t_i)-{(R\boldsymbol\beta)}_i\right]}^2,
    \quad |\beta_j|\leq0.1,
    \quad \sum_j|\beta_{j+1}-\beta_j|\leq0.1.
    \label{eq:lag-fit}
\end{equation}
Each fit uses 256 paired calibration trajectories and no image metric.

To isolate schedule shape, we compare a fitted schedule with a constant at the same step-weighted mean gain, 1.07216, at NFE 25.
On 50,000 exactly paired unguided SiT-XL/2 ImageNet-256 images per arm, FID is 13.0 for the schedule and 12.5 for the constant.
The schedule-minus-constant difference is $+0.5\;[+0.4,+0.6]$, with a paired 95\% interval from 2,000 common-row influence resamples.

\begin{table}[ht]
    \centering
    \setlength{\tabcolsep}{7pt}
    \caption{
        Population time lag RMS for unguided SiT-XL/2 on ImageNet-256 under uniform stochastic Euler sampling.
        Values are the RMS across saved times of the genuine-minus-generated mean raw-state probe reading, on the unit time scale; lower is better.
        Each schedule uses 256 calibration trajectories and is evaluated on 1,280 disjoint paths paired across interventions by class label, initial latent, and Brownian path.
        The direct scalar is selected and evaluated on those same 1,280 paths.
        The constant gains are the direct scalar lag minima, $\gamma_{\rm lag}$, from Table~\ref{tab:scalar-lag-calibration}.
        This comparison changes both average gain and schedule shape; the matched-average FID comparison above isolates shape.
        The table reports point estimates.
    }%
    \label{tab:lag-schedule-average}
    \begin{tabular}{@{}rrrr@{}}
        \toprule
        NFE & Unscaled & \shortstack{Fitted\\schedule} & \shortstack{Direct\\scalar} \\
        \midrule
        25  & 0.0254 & 0.0046 & 0.0058 \\
        50  & 0.0121 & 0.0020 & 0.0039 \\
        100 & 0.0061 & 0.0026 & 0.0026 \\
        250 & 0.0037 & 0.0023 & 0.0029 \\
        \bottomrule
    \end{tabular}
\end{table}
\FloatBarrier{}
\Needspace{12\baselineskip}
\section{Controls for alternative explanations}%
\label{app:controls}

These controls ask whether gain helps because the learned field is uniformly too small, the numerical solver is inaccurate, or the endpoint has the wrong basic moments.
The results do not support any of these as a full explanation.
Better integration still reduces population time lag and the lowest-FID gain, so numerical error contributes to the effect at low NFE without explaining all of it.
The Scale Schedule Corrector (SSC) of \citet{li2026velocitydeficit} uses the gain $\gamma(t)=1.1-0.1t$.

\Needspace{8\baselineskip}
\subsection{The trained field is not systematically underscaled}
\label{app:trained-field-scaling}

We first test the direct prediction that held-out target velocities should favor multiplying the learned field by more than one.

\begin{figure}[ht]
    \centering
    \includegraphics[width=0.82\textwidth]{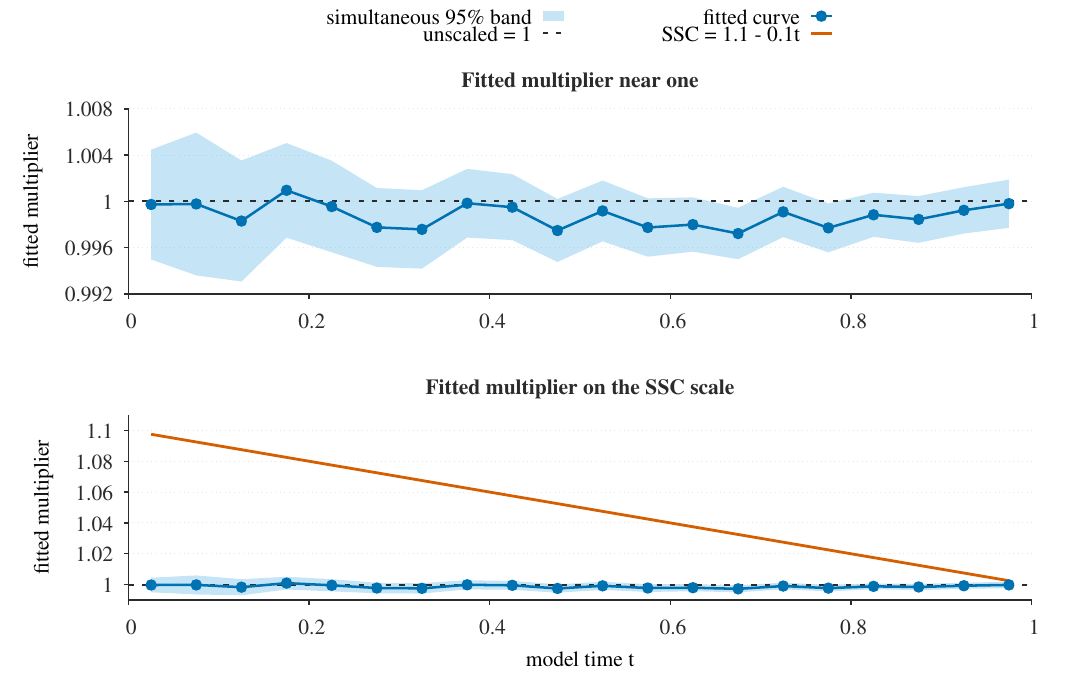}
    \caption{
        Held-out scalar multiplier fitted to the pretrained SiT-XL/2 velocity at 20 model times using 16,000 ImageNet image--noise pairs.
        The top panel expands the range around one; the bottom panel uses the scale that separates the fitted curve from SSC\@.
        The curve has mean 0.999, and the shaded region is a simultaneous 95\% band.
    }%
    \label{fig:fitted-gamma}
\end{figure}
\FloatBarrier{}

The fitted multiplier instead stays near one throughout the path.
The trained SiT field therefore does not show the systematic underscaling assumed by the velocity-deficit account.

\Needspace{12\baselineskip}
\subsection{Velocity scaling is not an Euler endpoint-error correction}

We next test whether SSC compensates for error from coarse Euler integration by comparing its endpoint with a fine unscaled reference.
The ODE reference uses adaptive Dormand--Prince 5(4) with relative tolerance $10^{-6}$ and absolute tolerance $10^{-7}$.
The SDE reference uses 1,000 stochastic-Heun steps, requiring 2,000 model evaluations.
Coarse and reference SDE trajectories share the same nested Brownian path.

\begin{table}[ht]
    \centering
    \small
    \setlength{\tabcolsep}{8pt}
    \caption{
        Endpoint error under solver refinement for SiT-XL/2 with Euler integration.
        $E_N$ is mean squared latent-coordinate endpoint error relative to the fine unscaled reference on 1,024 matched trajectories; lower is closer, and brackets are paired bootstrap 95\% intervals.
    }%
    \label{tab:solver-refinement}
    \begin{tabular}{@{}lrrr@{}}
        \toprule
        Dynamics & NFE & Unscaled $E_N$ [95\%]  & SSC $E_N$ [95\%]       \\
        \midrule
        ODE      & 25  & $0.059\ [0.055,0.064]$ & $0.071\ [0.068,0.074]$ \\
        ODE      & 250 & $0.004\ [0.004,0.005]$ & $0.052\ [0.050,0.055]$ \\
        SDE      & 25  & $0.096\ [0.090,0.101]$ & $0.067\ [0.063,0.072]$ \\
        SDE      & 250 & $0.008\ [0.007,0.009]$ & $0.065\ [0.063,0.068]$ \\
        \bottomrule
    \end{tabular}
\end{table}

Refining Euler integration sharply reduces unscaled endpoint error, while the SSC endpoint remains far from the fine unscaled reference.

We also isolate one-step error by restarting each NFE-50 Euler step from the fine unscaled reference state and comparing its result with the reference at the next time.
The SDE steps share the reference Brownian increment.
For comparison, we choose the SDE gain that minimizes one-step error in each of 20 time bins on separate calibration paths, then interpolate between these values.

\begin{table}[ht]
    \centering
    \small
    \setlength{\tabcolsep}{5pt}
    \caption{
        Isolated one-step numerical error for unguided SiT-XL/2 at NFE 50.
        $e$ is squared latent-coordinate error averaged over all 50 steps and 1,024 matched validation trajectories; lower is closer to the fine unscaled reference.
        The change is $10^6(e_{\rm scaled}-e_{\rm unscaled})$, with negative values indicating lower error.
        Brackets give paired 95\% intervals from 2,000 whole-trajectory bootstrap resamples.
    }%
    \label{tab:one-step-error}
    \begin{tabular}{@{}llrrr@{}}
        \toprule
        Dynamics & Gain      & Unscaled $e$          & Scaled $e$            & Change [95\%]            \\
        \midrule
        ODE      & SSC       & $3.58\times10^{-7}$ & $1.83\times10^{-6}$ & $+1.48\ [+1.46,+1.49]$ \\
        SDE      & SSC       & $4.10\times10^{-5}$ & $4.45\times10^{-5}$ & $+3.56\ [+3.53,+3.60]$ \\
        SDE      & Fitted    & $4.10\times10^{-5}$ & $4.06\times10^{-5}$ & $-0.33\ [-0.34,-0.32]$ \\
        \bottomrule
    \end{tabular}
\end{table}

SSC increases one-step error in both ODE and SDE sampling, whereas the fitted SDE schedule reduces it (Table~\ref{tab:one-step-error}).
Together with solver refinement, this shows that SSC changes the sampled path rather than simply correcting Euler error.

\Needspace{8\baselineskip}
\subsection{Heun reduces lag and the lowest-FID gain}

A higher-order solver tests whether better integration reduces both population time lag and the useful amount of gain at the same model-call budget.

\begin{table}[ht]
    \centering
    \small
    \caption{
        Matched-budget solver diagnostics on 1,280 exactly paired unguided SDE trajectories.
        Euler uses 50 steps and stochastic Heun uses 25 predictor--corrector steps, for 50 network evaluations with both solvers.
        Both endpoints are compared with the same-path 1,000-step stochastic-Heun reference.
        Lower is better for every row.
        The contrast is Heun minus Euler; brackets give paired 95\% trajectory-bootstrap intervals.
        Strong endpoint error is root-mean-square error in latent units per coordinate, and population time lag is measured on the unit time scale.
        Population time lag RMS is the root mean square over saved times of the population-mean probe mismatch.
    }%
    \label{tab:heun-diagnostics}
    \begin{tabular}{@{}L{5.1cm}rrrr@{}}
        \toprule
        Quantity                & Euler & Heun  & Contrast & 95\% interval     \\
        \midrule
        Strong endpoint error   & 0.208 & 0.148 & $-0.060$ & $[-0.066,-0.055]$ \\
        Population time lag RMS & 0.019 & 0.005 & $-0.014$ & $[-0.014,-0.013]$ \\
        \bottomrule
    \end{tabular}
\end{table}

At the same budget of 50 model evaluations, Heun reduces strong endpoint error from 0.208 to 0.148 and population time lag RMS from 0.019 to 0.005.
The lowest-FID gain also moves from 1.10 with Euler to 1.05 with Heun (Figure~\ref{fig:context-dependence}b).
Numerical integration therefore affects how much gain helps, even though the preceding control shows that SSC is not simply an endpoint-error correction.

\FloatBarrier{}
\Needspace{8\baselineskip}
\subsection{Matching endpoint moments does not reproduce the FID improvement}

All three controls act on completed endpoints before decoding.
The radius control uses one scalar fitted to match encoded real-data radius.
The directional control rescales centered coordinates to impose an idealized SSC-like variance pattern.
An affine map matches the full latent mean and covariance of gain-1.05 endpoints.

\begin{table}[ht]
    \centering
    \small
    \setlength{\tabcolsep}{2pt}
    \caption{
        FID for velocity scaling during sampling and endpoint-only controls on 50,000 ImageNet-256 images per intervention; lower is better.
        Rows with the same paired-set letter use the same images.
        Paired set A uses the public hybrid SDE sampler at NFE 50; paired sets B and C use uniform stochastic Euler sampling at NFE 100.
        The contrast is unscaled FID minus intervention FID, so a positive value favors the intervention; brackets give paired 95\% intervals.
        Paired set A intervals use 2,000 common-row first-order FID influence resamples with the real reference fixed; $-0.0$ denotes a negative value smaller than 0.05 in magnitude.
    }%
    \label{tab:endpoint-controls}
            \begin{tabular}{@{}L{2.7cm}ccL{2.6cm}rrL{3.1cm}@{}}
            \toprule
            Endpoint property & \shortstack{Paired\\set} & NFE & Intervention & \shortstack{Unscaled\\FID} & \shortstack{Intervention\\FID} & \shortstack{FID reduction\\{[95\%]}} \\
            \midrule
            Global radius                     & A          & 50  & Velocity scaling during sampling & 13.8         & 7.6              & $+6.2\;[+6.1,+6.4]$                \\
                                              & A          & 50  & Endpoint-only control            & 13.8         & 13.8             & $-0.0\;[-0.1,-0.0]$                \\
            \addlinespace
            Variance in 256 directions        & B          & 100 & Velocity scaling during sampling & 11.5         & 6.8              & $+4.7\;[+4.6,+4.9]$                \\
                                              & C          & 100 & Endpoint-only control            & 11.6         & 11.6             & $+0.0\;[0.0,+0.1]$                 \\
            \addlinespace
            Global latent mean and covariance & C          & 100 & Velocity scaling during sampling & 11.6         & 6.8              & $+4.8\;[+4.7,+4.9]$                \\
                                              & C          & 100 & Endpoint-only control            & 11.6         & 12.3             & $-0.7\;[-0.8,-0.7]$                \\
            \bottomrule
        \end{tabular}
\end{table}

\FloatBarrier{}
Matching radius or directional variance after sampling leaves FID essentially unchanged.
Matching the full latent mean and covariance worsens FID, whereas applying gain during sampling lowers it.
These global moment changes do not reproduce velocity scaling along the stochastic sampling path.

Together, the controls do not support uniform field underscaling, simple Euler-error correction, or endpoint-moment matching as full explanations for why gain improves image quality.

\FloatBarrier{}
\Needspace{10\baselineskip}
\section{Changed-state response}%
\label{app:response}

\subsection{Pathwise decomposition}
\label{app:path-decomposition}

For the SiT SDE drift used by both the uniform and hybrid samplers, the gain-$\gamma$ drift is $b_\gamma(x,t)=(1+t)\gamma v(x,t)-x$.
Let $x_k$ be the unscaled state and $y_k$ the constant-gain state at step $k$, driven by the same Brownian increment $\eta_k$, and define $\Delta_k=y_k-x_k$.
One stochastic Euler step gives
\begin{equation}
    \begin{aligned}
        x_{k+1}
         & =x_k+h_k\big[(1+t_k)v(x_k,t_k)-x_k\big]+\eta_k,        \\
        y_{k+1}
         & =y_k+h_k\big[(1+t_k)\gamma v(y_k,t_k)-y_k\big]+\eta_k, \\
        \Delta_{k+1}
         & =(1-h_k)\Delta_k
        +h_k(1+t_k)(\gamma-1)v(x_k,t_k)                           \\
         & \quad+h_k(1+t_k)\gamma\big[v(y_k,t_k)-v(x_k,t_k)\big].
    \end{aligned}
    \label{eq:path-decomposition}
\end{equation}
For the hybrid sampler, the same decomposition holds for the final Mean update, where $\eta_k=0$.
The term proportional to $\gamma-1$ is the direct push retained by the frozen-field control.
The final term is the changed-state response removed by that control.

\begin{table}[ht]
    \centering
    \small
    \setlength{\tabcolsep}{4pt}
    \caption{
        Frozen-field FID reduction across samplers, SiT checkpoints, and an ADM U-Net.
        Each intervention uses 50,000 ImageNet-256 images.
        Within each row, corresponding images share the class label and initial latent; stochastic rows also share the Brownian path.
        The FID column gives values as unscaled $\rightarrow$ frozen field $\rightarrow$ full scaling.
        The FID-reduction ratio is $(\mathrm{FID}_{\rm unscaled}-\mathrm{FID}_{\rm frozen})/(\mathrm{FID}_{\rm unscaled}-\mathrm{FID}_{\rm full})$.
        Ratios use unrounded FIDs; displayed FIDs have one decimal place.
        Brackets give paired 95\% intervals for the ratio within each sampler.
    }%
    \label{tab:response-transfer}
    \begin{tabular}{@{}L{2.4cm}L{2.3cm}cL{3.1cm}@{}}
        \toprule
        Model & Sampler, NFE, gain & \shortstack{FID\@: unscaled\\$\rightarrow$ frozen $\rightarrow$ full} & \shortstack{FID-reduction\\ratio [95\%]} \\
        \midrule
        SiT-S/2       & Uniform stochastic Euler, 25, 1.07 & $71.3\rightarrow66.3\rightarrow51.3$                    & 24.9\% $[24.4,25.4]$       \\
        SiT-B/2       & Uniform stochastic Euler, 25, 1.07 & $47.9\rightarrow43.7\rightarrow30.3$                    & 23.8\% $[23.4,24.3]$       \\
        SiT-L/2       & Uniform stochastic Euler, 25, 1.07 & $30.2\rightarrow26.8\rightarrow16.1$                    & 23.6\% $[23.1,24.0]$       \\
        SiT-XL/2      & Public hybrid SDE, 50, 1.05        & $13.8\rightarrow12.4\rightarrow7.7$                     & 22.4\% $[21.8,23.0]$       \\
        REPA-SiT-XL/2 & Public hybrid SDE, 50, 1.05        & $10.3\rightarrow9.2\rightarrow5.3$                      & 22.7\% $[22.1,23.4]$       \\
        \addlinespace
        SiT-XL/2      & ODE, 50, 1.05                      & $11.2\rightarrow9.7\rightarrow8.5$                      & 53.9\% $[51.4,56.4]$       \\
        LFM ADM U-Net & ODE, 50, 1.05                      & $19.6\rightarrow17.4\rightarrow15.9$                    & 59.0\% $[56.6,61.4]$       \\
        \bottomrule
    \end{tabular}
\end{table}
\FloatBarrier{}

\Needspace{8\baselineskip}
\subsection{A constant gain propagates differently under ODE and SDE sampling}

To show how sampler dynamics change the frozen-field push, we examine the continuous ODE and SDE corresponding to the discrete updates.
Freeze the neural field to the unscaled path $x(t)$.
Let $z(t)$ be the frozen-field path and define $\Delta_{\mathrm{fr}}(t)=z(t)-x(t)$ with $\Delta_{\mathrm{fr}}(0)=0$.

Under ODE sampling, the frozen-field difference satisfies
\begin{equation}
    \dot{\Delta}_{\mathrm{fr}}(t)
    =(\gamma-1)v(x(t),t).
\end{equation}
Since the unscaled path obeys $\dot{x}(t)=v(x(t),t)$,
\begin{equation}
    \Delta_{\mathrm{fr}}(t)
    =(\gamma-1)\int_0^t v(x(s),s)\,ds
    =(\gamma-1)\bigl(x(t)-x(0)\bigr).
    \label{eq:frozen-ode-closed-form}
\end{equation}
All intermediate velocity terms telescope.
The frozen-field path is therefore the per-sample affine map
\begin{equation}
    z(t)=\gamma x(t)+(1-\gamma)x(0).
\end{equation}
The frozen-field ODE control therefore needs no additional model calls: after unscaled sampling, we extrapolate each endpoint along its displacement from the initial noise.
At NFE 50, this update lowers FID from 11.2 to 9.7.

For the SDE drift above, the shared Brownian path cancels from the frozen-field difference.
The remaining dynamics are
\begin{equation}
    \dot{\Delta}_{\mathrm{fr}}(t)
    =(\gamma-1)(1+t)v(x(t),t)-\Delta_{\mathrm{fr}}(t).
\end{equation}
Solving this linear equation gives
\begin{equation}
    \Delta_{\mathrm{fr}}(t)
    =(\gamma-1)\int_0^t e^{-(t-s)}(1+s)v(x(s),s)\,ds.
    \label{eq:frozen-sde-closed-form}
\end{equation}
At the endpoint, the velocity at time $s$ receives weight
\begin{equation}
    w(s)=e^{-(1-s)}(1+s).
\end{equation}
This weight rises smoothly across the path while preserving unit total weight:
\begin{equation}
    w(0)=e^{-1},
    \qquad
    w(1)=2,
    \qquad
    \int_0^1 w(s)\,ds=1.
\end{equation}
Under the frozen-field control, the ODE weights unscaled-path velocities uniformly across time.
The SDE gives later velocities more endpoint weight because an earlier displacement is reduced by the $-x$ drift for longer and the velocity coefficient $1+t$ grows across the path.
The gain remains constant; this weighting comes from the SDE dynamics.

\begin{minipage}{\textwidth}
    \subsection{Deterministic ODE gain curve}%
    \label{app:ode-gain-turn}

    Full-response FID falls from 11.2 at gain one to 8.5 at gain 1.05, then rises to 9.5 at gain 1.075 and continues to rise at every larger tested gain.
    The full-response 1.075-minus-1.05 contrast is $+1.0$ with a paired 95\% interval $[+0.9,+1.1]$.
    With the neural field frozen to the unscaled path, gains 1.05 and 1.075 both give FID 9.7.
    Model reevaluation therefore gives a large additional FID reduction at gain 1.05 but little additional reduction at gain 1.075.
    The changed-state response helps determine which gain works best, although this comparison does not explain why its benefit becomes smaller at the higher gain.

    \centering
    \captionsetup{type=table}
    \small
    \setlength{\tabcolsep}{3.8pt}
    \captionof{table}{
        Deterministic ODE gain scan for unguided SiT-XL/2 on ImageNet-256 under Euler sampling at NFE 50.
        Each gain uses 50,000 exactly paired images, and lower FID is better.
        For each row after unscaled, $\Delta\mathrm{FID}=\mathrm{FID}(\text{current gain})-\mathrm{FID}(\text{preceding gain})$, so a negative contrast favors the current gain.
        Contrasts are computed from unrounded FID values; displayed FIDs use one decimal place.
        Brackets give paired 95\% intervals from 2,000 common-row influence resamples with the real reference fixed.
        The lowest-FID gain is the tested gain with the lowest measured FID on this grid, not a continuous-optimum estimate.
    }%
    \label{tab:e92-ode-gain}
    \begin{tabular}{@{}rrrrrr@{}}
        \toprule
        Gain & FID $\downarrow$ & \shortstack{Preceding\\gain} & \shortstack{Preceding\\FID} & $\Delta$FID & \shortstack{Paired 95\%\\interval} \\
        \midrule
        1.000 & 11.2             & --             & --            & --          & --                   \\
        1.050 & 8.5              & 1.000          & 11.2          & $-2.6$      & $[-2.8,-2.5]$        \\
        1.075 & 9.5              & 1.050          & 8.5           & $+1.0$      & $[+0.9,+1.1]$        \\
        1.100 & 11.7             & 1.075          & 9.5           & $+2.2$      & $[+2.1,+2.3]$        \\
        1.125 & 15.0             & 1.100          & 11.7          & $+3.3$      & $[+3.2,+3.4]$        \\
        1.150 & 18.9             & 1.125          & 15.0          & $+3.9$      & $[+3.8,+4.0]$        \\
        1.175 & 23.6             & 1.150          & 18.9          & $+4.6$      & $[+4.5,+4.7]$        \\
        1.200 & 28.5             & 1.175          & 23.6          & $+4.9$      & $[+4.8,+5.0]$        \\
        1.225 & 33.7             & 1.200          & 28.5          & $+5.2$      & $[+5.0,+5.3]$        \\
        \bottomrule
    \end{tabular}
\end{minipage}

\Needspace{8\baselineskip}
\subsection{The realized response direction matters}
\label{app:response-direction}

\begin{figure}[ht]
    \centering
    \includegraphics[width=\textwidth,trim=0 16bp 0 0,clip]{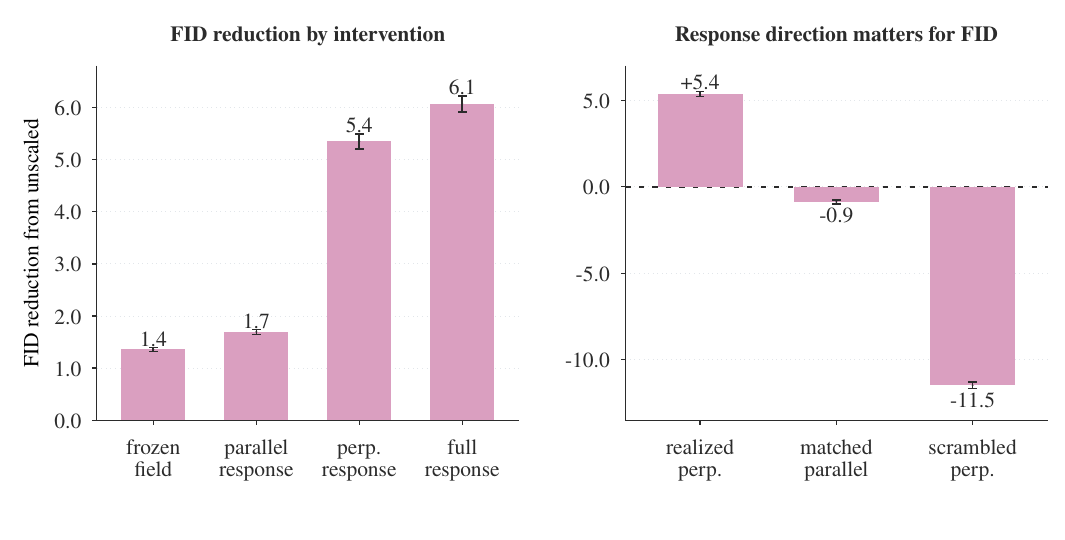}
    \caption{
        Changed-state response components and direction controls for unguided SiT-XL/2 on ImageNet-256 under public hybrid SDE sampling at NFE 50 and gain 1.05.
        The vertical quantity is the FID reduction from unscaled sampling in FID points, so higher is better; unscaled FID is 13.8.
        Left: each bar shows a separate intervention, not an additive component.
        Right: equal-norm parallel and scrambled perpendicular replacements raise FID\@.
        Each panel uses a separate population of 50,000 exactly paired images; error bars are paired 95\% intervals from 2,000 common-row influence resamples with the real reference fixed.
        Bar heights and contrasts use unrounded FIDs; raw FIDs elsewhere are displayed to one decimal.
        Table~\ref{tab:response-direction} gives raw values and exact intervals for the direction controls.
    }%
    \label{fig:response-components}
\end{figure}

\paragraph{Changed-state response components.}
We compare interventions that retain only the parallel or perpendicular component of the changed-state response.
Writing this response as $r_k=v_\theta(y_k,t_k)-v_\theta(x_k,t_k)$, we decompose it as $r_k=r_k^\parallel+r_k^\perp$, where the two components are parallel and perpendicular to the unscaled velocity $v_\theta(x_k,t_k)$.
Let $y_k^a$ be the current state of component intervention $a\in\{\parallel,\perp\}$.
At each step, its scaled velocity is
\begin{equation}
    \widetilde v_k^a
    =\gamma_k\left(
    v_\theta(x_k,t_k)
    +P_k^a\!\left[v_\theta(y_k^a,t_k)-v_\theta(x_k,t_k)\right]
    \right),
    \label{eq:component-intervention}
\end{equation}
where $P_k^a$ selects the component parallel or perpendicular to the unscaled velocity $v_\theta(x_k,t_k)$.
The sampler uses $\widetilde v_k^a$ in its velocity term.
For unguided SiT-XL/2 with the public NFE-50 hybrid SDE sampler and gain 1.05, the perpendicular-only intervention gives FID 8.4, compared with 12.1 for the parallel-only intervention and 7.7 for full scaling (Figure~\ref{fig:response-components}).

\paragraph{Changed-state response direction.}
Because the parallel and perpendicular components differ in norm, the component comparison does not isolate direction from magnitude.
We therefore test two controls that replace the perpendicular response while preserving its norm.
Let $x_k$ and $y_k^\perp$ denote the states of the unscaled and perpendicular-only reference trajectories.
At each step, the realized perpendicular response is the component of $v(y_k^\perp,t_k)-v(x_k,t_k)$ orthogonal to the unscaled velocity $v(x_k,t_k)$.
The parallel control replaces this component with an equal-norm vector in the positive unscaled-velocity direction.
The scrambled control applies one trajectory-specific signed coordinate permutation, fixed across time, to the realized perpendicular response; at each step, it projects the result perpendicular to the unscaled velocity and restores the original response norm.
Both replacement controls retain the direct gain push and use reoriented copies of this reference response.
We re-evaluate the unscaled and perpendicular-only reference states at every step, but do not evaluate the model at the replacement controls' states.

\begin{table}[ht]
    \centering
    \small
    \setlength{\tabcolsep}{4pt}
    \caption{
        Response-direction controls for unguided SiT-XL/2 on ImageNet-256 under stochastic sampling.
        All three blocks use hybrid SDE sampling at constant gain 1.05.
        The first two use NFE 50; the last uses NFE 25 and 250.
        Each intervention uses 50,000 images, and lower FID and DINOv2 feature distance are better.
        Within each comparison, the two settings have exact label, initial-latent, Brownian-stream, and image-row pairing.
        The controls match the norm of the realized perpendicular velocity change at every step but change its direction.
        Each contrast is realized perpendicular minus comparator, so negative favors the realized response.
        Raw DINOv2 values are Fr\'echet distances in squared feature units and are rounded to whole units; contrasts use unrounded metric values.
        Brackets give paired 95\% intervals from 2,000 common-row influence resamples with the real reference fixed.
    }%
    \label{tab:response-direction}
    \begin{tabular}{@{}L{3.0cm}rrl@{}}
        \toprule
        Comparator & \shortstack{Realized\\$\perp$ FID} & \shortstack{Comparator\\FID} & FID contrast [95\%]    \\
        \midrule
        Unscaled                 & 8.5                  & 13.8           & $-5.4\;[-5.5,-5.2]$    \\
        Norm-matched $\parallel$ & 8.5                  & 14.7           & $-6.2\;[-6.4,-6.1]$    \\
        Scrambled $\perp$        & 8.5                  & 25.3           & $-16.8\;[-17.1,-16.6]$ \\
        \bottomrule
    \end{tabular}
    \par\vspace{0.5\baselineskip}
    \begin{tabular}{@{}L{3.0cm}rrl@{}}
        \toprule
        Comparator & \shortstack{Realized\\$\perp$ DINO} & \shortstack{Comparator\\DINO} & DINO contrast [95\%]      \\
        \midrule
        Unscaled                 & 269                   & 340             & $-71.5\;[-72.8,-70.3]$    \\
        Norm-matched $\parallel$ & 269                   & 316             & $-47.7\;[-48.8,-46.5]$    \\
        Scrambled $\perp$        & 269                   & 440             & $-171.5\;[-173.5,-169.7]$ \\
        \bottomrule
    \end{tabular}
    \par\vspace{0.5\baselineskip}
    \begin{tabular}{@{}crrl@{}}
        \toprule
        NFE & \shortstack{Realized\\$\perp$ FID} & \shortstack{Equal-norm\\$\parallel$ FID} & FID contrast [95\%] \\
        \midrule
        25  & 16.3                 & 21.2                       & $-4.9\;[-5.0,-4.7]$ \\
        250 & 6.5                  & 12.5                       & $-6.0\;[-6.2,-5.9]$ \\
        \bottomrule
    \end{tabular}
\end{table}

At NFE 50, the realized perpendicular response has lower FID and lower DINOv2 feature distance than each equal-norm direction control.
The same FID ordering at NFE 25 and 250 shows that the response direction matters beyond its norm.

\FloatBarrier{}
\Needspace{8\baselineskip}
\subsection{Direction of the direct extra push}
\label{app:direct-push-direction}

We compare unscaled sampling with velocity scaling and the radial control at gain $\gamma=1.05$, using unguided SiT-XL/2 on ImageNet-256 under public hybrid SDE sampling at NFE 50.
Each intervention uses 50,000 images paired by class label, initial latent, and Brownian path.
At any given state, velocity scaling and the radial control add pushes with equal norms; their magnitudes can differ after their trajectories diverge.

Velocity scaling lowers FID from 13.9 to 7.8, while the radial control raises it to 19.4.
Relative to unscaled sampling, the FID change is $-6.0\;[-6.2,-5.9]$ for velocity scaling and $+5.5\;[+5.4,+5.7]$ for the radial control.
These contrasts subtract unscaled FID and use unrounded values.
Brackets give paired 95\% intervals from 2,000 common-row influence resamples with the real reference fixed.

\FloatBarrier{}
\section{Distributional effects of gain}%
\label{app:gain}

Population time lag explains only part of the FID improvement.
We therefore ask what else gain changes in the generated distribution.
We first show that gain trades fidelity against diversity and redistributes endpoint variance across real-data covariance directions.
We also relate this variance redistribution to the extra gain beyond lag correction.
We then test and reject a simpler account in which gain raises the whole endpoint density to one shared power.

\subsection{Image-metric tradeoff}

We first measure how the image metrics change as constant gain increases.

\begin{table}[ht]
    \centering
    \small
    \caption{
        Image metrics at four constant gains for unguided SiT-XL/2 with uniform stochastic Euler sampling at NFE 25 and 50,000 exactly paired ImageNet-256 images per gain.
        Gain 1.000 is unscaled and gain 1.150 is the lowest-FID gain on the tested grid.
        Each contrast is the value at the higher gain minus the value at the lower gain and is computed from unrounded metric values.
        Lower FID and sFID are better; higher precision and recall are better.
        Brackets are paired 95\% intervals from 2,000 common-row first-order influence resamples for FID and sFID or a 20-group delete-one jackknife for precision and recall.
    }%
    \label{tab:nfe25-metric-tradeoff}
    \setlength{\tabcolsep}{7pt}
    \begin{tabular}{@{}lrrrr@{}}
        \toprule
        Gain                    & FID $\downarrow$ & sFID $\downarrow$ & Precision $\uparrow$ & Recall $\uparrow$ \\
        \midrule
        1.000 (unscaled)        & 28.0             & 26.2              & 0.565                & 0.630             \\
        1.072  & 12.7             & 8.3               & 0.653                & 0.649             \\
        1.100                   & 10.0             & 5.6               & 0.675                & 0.639             \\
        1.150 (lowest-FID gain) & 8.7              & 6.7               & 0.685                & 0.622             \\
        \bottomrule
    \end{tabular}
    \par\vspace{0.5\baselineskip}
    \setlength{\tabcolsep}{3.5pt}
            \begin{tabular}{@{}lrr@{}}
\toprule
Gain contrast & $\Delta$FID [95\%] & $\Delta$sFID [95\%] \\
\midrule
$1.100-1.072$ & $-2.7\;[-2.8,-2.6]$ & $-2.7\;[-2.8,-2.6]$ \\
$1.150-1.100$ & $-1.3\;[-1.4,-1.2]$ & $+1.1\;[+1.0,+1.2]$ \\
\bottomrule
\end{tabular}
\par\vspace{0.7\baselineskip}
\begin{tabular}{@{}lrr@{}}
\toprule
Gain contrast & $\Delta$precision [95\%] & $\Delta$recall [95\%] \\
\midrule
$1.100-1.072$ & $+0.022\;[+0.019,+0.025]$ & $-0.009\;[-0.023,+0.005]$ \\
$1.150-1.100$ & $+0.010\;[+0.006,+0.014]$ & $-0.018\;[-0.031,-0.004]$ \\
\bottomrule
\end{tabular}
\end{table}

Increasing gain from 1.072 to 1.100 improves FID, sFID, and precision, while recall changes little within uncertainty.
Increasing gain to the lowest-FID value of 1.150 improves FID and precision further but worsens sFID and recall.
Gain therefore creates a trade-off rather than improving every metric together.

\Needspace{16\baselineskip}
\subsection{Scaling sharpens the endpoint covariance spectrum}%
\label{app:spectral-sharpening}

We measure how gain redistributes endpoint variance under stochastic sampling.
The measurement uses 256 frozen, cross-fitted real-data directions.
For each sampling setting, the \emph{power slope} $b$ is the ordinary-least-squares coefficient in
\begin{equation}
    \log \operatorname{Var}_{\rm gen,j}=a+b\log \operatorname{Var}_{\rm data,j},
\end{equation}
across directions $j$.
The relative power slope $R=b_{1.05}/b_1$ compares constant gain 1.05 with unscaled sampling; $R>1$ means that scaling moves relatively more endpoint variance toward directions where held-out real data vary more.

\begin{table}[!ht]
    \centering
    \small
    \setlength{\tabcolsep}{4pt}
    \caption{
    Relative endpoint power slope for unguided SiT-XL/2 and a matched linear Gaussian SDE\@; neither larger nor smaller is intrinsically better for image quality.
    $R=b_{1.05}/b_1$ is dimensionless, and $R>1$ means more variance toward directions with larger held-out real-data variance under gain 1.05 than under unscaled sampling.
    The signed contrast is SiT $R$ minus Gaussian $R$.
    Euler--Maruyama rows use 1,280 exactly paired endpoints per setting; the separate 1,000-step stochastic-Heun reference (NFE 2,000) uses 1,024 endpoints per setting, and every row uses 1,024 real-data latents over 256 fixed cross-fitted directions.
    Brackets give 95\% intervals from 2,000 paired fixed-basis bootstrap draws.
    }%
    \label{tab:spectral-sharpening}
            \begin{tabular}{@{}lrrr@{}}
\toprule
Solver & NFE & SiT $R$ [95\%] & Gaussian $R$ [95\%] \\
\midrule
Euler--Maruyama & 25 & $1.0628\;[1.0612,1.0643]$ & $1.0527\;[1.0526,1.0528]$ \\
Euler--Maruyama & 50 & $1.0607\;[1.0590,1.0623]$ & $1.0543\;[1.0542,1.0543]$ \\
Euler--Maruyama & 100 & $1.0595\;[1.0577,1.0612]$ & $1.0551\;[1.0550,1.0551]$ \\
Euler--Maruyama & 250 & $1.0574\;[1.0556,1.0594]$ & $1.0556\;[1.0555,1.0556]$ \\
Stochastic Heun & 2000 & $1.0563\;[1.0543,1.0583]$ & $1.0559\;[1.0558,1.0559]$ \\
\bottomrule
\end{tabular}
\par\vspace{0.7\baselineskip}
\begin{tabular}{@{}lrr@{}}
\toprule
Solver & NFE & SiT minus Gaussian [95\%] \\
\midrule
Euler--Maruyama & 25 & $+0.0101\;[+0.0085,+0.0116]$ \\
Euler--Maruyama & 50 & $+0.0064\;[+0.0048,+0.0081]$ \\
Euler--Maruyama & 100 & $+0.0044\;[+0.0026,+0.0062]$ \\
Euler--Maruyama & 250 & $+0.0018\;[+0.0001,+0.0039]$ \\
Stochastic Heun & 2000 & $+0.0004\;[-0.0015,+0.0024]$ \\
\bottomrule
\end{tabular}
\end{table}

\Needspace{7\baselineskip}
The SiT relative power slope is about 1.06 at every tested NFE, so scaling sharpens the endpoint covariance spectrum under stochastic sampling.
The SiT--Gaussian difference shrinks with solver refinement and is $+0.0004\ [-0.0015,+0.0024]$ for the 1,000-step stochastic-Heun reference, where the two responses agree within uncertainty.
The Gaussian model does not match the full learned endpoint covariance, and this measurement does not show that this sharpening causes the FID reduction.

\Needspace{12\baselineskip}
\paragraph{Sharpening beyond lag correction.}
For unguided SiT-XL/2, we measure sharpening relative to the lag-based sampler.
Let $b_N(\gamma)$ denote the power slope at gain $\gamma$ and NFE $N$.
The relative sharpening is
\begin{equation}
    S_N(\gamma)=\frac{b_N(\gamma)}{b_N(\gamma_{\rm lag}(N))}.
    \label{eq:sharpening-beyond-lag}
\end{equation}

At NFE 25, multiplying the lag-based gain by $\kappa$ gives the best tested FID gain.
We measure the relative sharpening at this gain and use it as our target:
\begin{equation}
    S^\star=S_{25}\!\left(\kappa\gamma_{\rm lag}(25)\right).
\end{equation}

\begin{figure}[!ht]
    \centering
    \includegraphics[width=\textwidth]{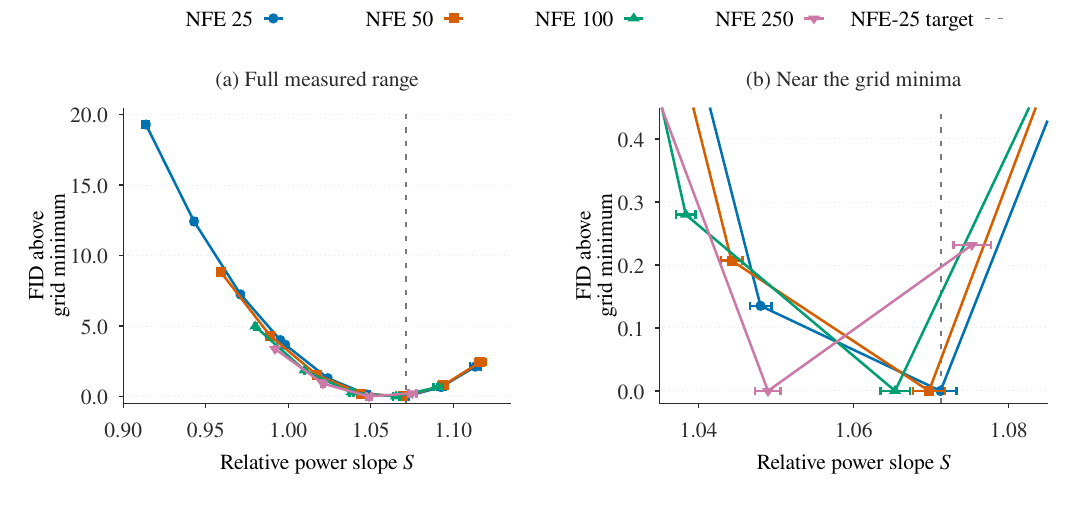}
    \caption{
        FID and covariance sharpening beyond lag correction for unguided SiT-XL/2 with uniform stochastic Euler sampling.
        Each point shows measured FID minus the lowest measured FID at the same NFE; lower is better.
        The horizontal coordinate is the relative power slope $S_N(\gamma)$ in Equation~\ref{eq:sharpening-beyond-lag}.
        Segments join measured gain settings, and the dashed line marks the target fitted at the best tested NFE-25 gain.
        Panel (b) enlarges the region near the grid minima.
        Each FID uses 50,000 images, and each power slope uses 1,280 paired endpoints in 256 cross-fitted directions from 1,024 real-data latents.
        We estimate the gains that match the sharpening target, and their FIDs, by linear interpolation between tested settings.
        Horizontal bars show paired 95\% intervals from 2,000 trajectory resamples, conditional on the real-data directions, lag-based gains, and FID grids.
        They do not include FID uncertainty or uncertainty in locating the best gain between tested settings.
    }%
    \label{fig:sharpening-transfer}
\end{figure}

At each higher NFE, we find the gain $\gamma_{\rm sharp}(N)$ that gives the same relative sharpening.
Its extra multiplier is
\begin{equation}
    \kappa_{\rm sharp}(N)=\frac{\gamma_{\rm sharp}(N)}{\gamma_{\rm lag}(N)}.
\end{equation}
These multipliers remain close to the NFE-25 value of $\kappa$.
Their estimated FIDs are within 0.2 of the lowest measured values at NFE 50, 100, and 250 (Figure~\ref{fig:sharpening-transfer}).

\FloatBarrier{}
\Needspace{8\baselineskip}
\subsection{One density-power coefficient does not describe scaling}%
\label{app:temperature-test}

A single density exponent would describe the full distribution, not only its covariance.
We test this broader account on continuous-ODE endpoints, where we can estimate likelihoods.
If scaling applied one global power transformation to the endpoint density, the scaled and unscaled densities would satisfy
\begin{equation}
    p_{1.05}(x\mid y)\propto p_1{(x\mid y)}^\beta,
\end{equation}
with one coefficient $\beta$ shared by every sample and direction after allowing each class its own normalization.

Taking logs gives the testable law
\begin{equation}
    \ell_{1.05}(x)-\ell_1(x)
    \approx c_y+(\beta-1)\ell_1(x).
\end{equation}
Here $\ell_\gamma(x)=\log p_\gamma(x\mid y)$, and $c_y$ is a class offset.
The analysis uses 126 endpoints from the unscaled and gain-1.05 ODEs across 16 ImageNet classes, split equally between fitting and evaluation.
Backward ODE integration estimates both log densities using velocity divergence averaged over 16 random $\pm1$ vectors shared between gains.

The fitted law left essentially all held-out response variance unexplained.
The residual-variance ratio was $1.14\ [0.90,1.92]$, where zero denotes exact agreement and one means that the fit explains none of the response variance.
The model-normalized root-mean-square residual error was $1.07\ [0.95,1.39]$.
Brackets contain 95\% of the stratified bootstrap distribution, and smaller values mean closer agreement with one common density-power law.

A matched Gaussian control recovered its known coefficient to within 0.01\% and had at most 0.22\% error in the full predicted log-density response.
Trace-estimation and ODE-refinement errors were also much smaller than the neural endpoint residual.
These held-out results do not support one common density-power coefficient for the tested gain-1.05 ODE change.

\FloatBarrier{}

\Needspace{12\baselineskip}
\section{Transfer and context dependence}
\label{app:transfer}

Gain above one improves FID across the ImageNet SiT and ADM systems, but not in the decoder-free CIFAR-10 system.
The lowest-FID gain also changes with NFE and classifier-free guidance.
This section shows which findings transfer and why one fixed gain is not universal.

\Needspace{14\baselineskip}
\subsection{A decoder-free boundary case}

We next test whether gain above one transfers to a model that evolves pixels directly, without a decoder.
The decoder-free CIFAR-10 system does not benefit from gain above one.
The two deterministic curves and all four stochastic SDE curves have an interior minimum at gain 1.0, and every other tested gain has higher FID\@.
For the deterministic solvers, the paired intervals comparing each adjacent gain with identity exclude zero.
The Inception Score point estimate is highest at gain 0.975 under both solvers, although gain 0.975 and identity both round to 10 at whole-number display precision and no Inception Score interval was computed.

\begin{figure}[ht]
    \centering
    \includegraphics[width=\textwidth]{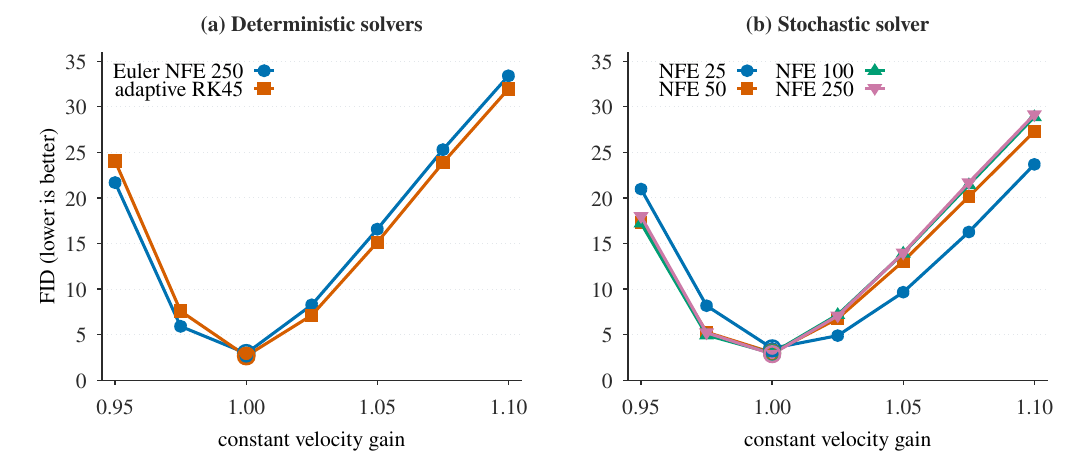}
    \caption{
        FID across constant gains for the official decoder-free CIFAR-10 1-Rectified Flow model; lower is better and open markers show the lowest measured FID\@.
        Panel (a) compares Euler NFE 250 with the official adaptive RK45 solver.
        Panel (b) uses the velocity-derived SDE in Equation~\ref{eq:velocity-derived-sde}.
        Each point uses 50,000 paired images within its solver and NFE\@.
        All six curves select gain one.
    }%
    \label{fig:decoder-free-gain}%
    \label{fig:decoder-free-sde-gain}
\end{figure}

\Needspace{30\baselineskip}
The figure shows the complete deterministic and stochastic gain curves.
Table~\ref{tab:decoder-free-gain} gives the exact deterministic FIDs and paired intervals for the gains next to identity.

\begin{table}[ht]
    \centering
    \small
    \setlength{\tabcolsep}{4.2pt}
    \caption{
        Decoder-free CIFAR-10 1-Rectified Flow gain curves under Euler NFE 250 and the official adaptive RK45 sampler.
        FID uses the official TF-Hub Inception network and CIFAR-10 reference; lower is better.
        The signed contrast is $\Delta\mathrm{FID}=\mathrm{FID}(\text{gain})-\mathrm{FID}(1)$, so a positive value favors identity.
        Contrasts are computed from unrounded FID values; displayed FIDs use one decimal place.
        Each gain setting contains 50,000 images.
        For the two gains neighboring identity, brackets give paired 95\% intervals from 2,000 common-row FID influence resamples with the real reference fixed.
    }%
    \label{tab:decoder-free-gain}
    \begin{tabular}{@{}lrrr@{}}
        \toprule
        Solver        & Gain  & FID $\downarrow$ & $\Delta$FID [95\%]  \\
        \midrule
        Euler NFE 250 & 0.950 & 21.7             & $+18.7$             \\
                      & 0.975 & 5.9              & $+3.0\;[+2.8,+3.1]$ \\
                      & 1.000 & 3.0              & $+0.0$              \\
                      & 1.025 & 8.3              & $+5.3\;[+5.2,+5.4]$ \\
                      & 1.050 & 16.6             & $+13.6$             \\
                      & 1.075 & 25.3             & $+22.3$             \\
                      & 1.100 & 33.4             & $+30.4$             \\
        \addlinespace
        Adaptive RK45 & 0.950 & 24.1             & $+21.4$             \\
                      & 0.975 & 7.6              & $+4.9\;[+4.8,+5.1]$ \\
                      & 1.000 & 2.7              & $+0.0$              \\
                      & 1.025 & 7.1              & $+4.5\;[+4.4,+4.6]$ \\
                      & 1.050 & 15.1             & $+12.5$             \\
                      & 1.075 & 23.9             & $+21.2$             \\
                      & 1.100 & 32.0             & $+29.3$             \\
        \bottomrule
    \end{tabular}
\end{table}
\FloatBarrier{}

\subsection{Guidance changes the lowest-FID gain}%
\label{app:cfg-gain}

The lowest-FID gain falls as classifier-free guidance grows, and velocity magnitude explains only part of this change.
We compare the velocity-magnitude prediction with changes in population time lag, endpoint covariance allocation, and image-metric trade-offs.

\Needspace{8\baselineskip}
\paragraph{Velocity-magnitude prediction.}
Stronger guidance increases velocity magnitude, so reducing gain could offset that increase.
We test whether this explains how much the lowest-FID gain falls as guidance grows.
Write the guided field as $v_\omega=v_c+(\omega-1)\Delta$, where $v_c$ is the conditional velocity and $\Delta=v_c-v_\varnothing$ is the guidance residual.
At a fixed state, the fractional change in velocity magnitude per unit guidance is
\begin{equation}
    \frac{\partial\log\lVert v_\omega\rVert}{\partial\omega}
    =
    \frac{\langle v_\omega,\Delta\rangle}{\lVert v_\omega\rVert^2}.
    \label{eq:cfg-magnitude-slope}
\end{equation}
We pooled this projection with the sampler step lengths over every model call along 1,024 gain-one trajectories at each guidance value.
If gain canceled only this change, each unit increase in guidance would lower the gain by about $c_{\rm pool}$.

\begin{table}[ht]
    \centering
    \small
    \setlength{\tabcolsep}{3.0pt}
    \caption{
    Fixed-state classifier-free-guidance field projection for SiT-XL/2 under the public hybrid SDE sampler at NFE 50 and gain one.
    Each row contains 1,024 trajectories.
    The pooled coefficient is $c_{\rm pool}=\sum_t \Delta t\langle v_\omega,\Delta\rangle/\sum_t \Delta t\lVert v_\omega\rVert^2$; a magnitude-only account predicts $d\gamma^\star/d\omega\approx-c_{\rm pool}$.
    The pointwise coefficient weights model calls equally; the lower block gives the pooled norm ratio and cosine that form the projection.
    No quantity has a favorable direction.
    Brackets give 95\% intervals from 2,000 whole-trajectory bootstrap resamples.
    }%
    \label{tab:cfg-gain-projection}
            \begin{tabular}{@{}rccc@{}}
\toprule
CFG $\omega$ & $c_{\rm pool}$ [95\%] & \shortstack{Pointwise coefficient\\{[95\%]}} & \shortstack{Lowest-FID\\gain} \\
\midrule
1.75 & 0.0258 [0.0252, 0.0264] & 0.0251 [0.0245, 0.0256] & 1.000 \\
2.00 & 0.0279 [0.0272, 0.0286] & 0.0271 [0.0265, 0.0276] & 0.975 \\
2.25 & 0.0298 [0.0292, 0.0305] & 0.0289 [0.0283, 0.0294] & 0.950 \\
\bottomrule
\end{tabular}
\par\vspace{0.7\baselineskip}
\begin{tabular}{@{}rcc@{}}
\toprule
CFG $\omega$ & $\lVert\Delta\rVert/\lVert v_\omega\rVert$ [95\%] & $\cos(\Delta,v_\omega)$ [95\%] \\
\midrule
1.75 & 0.119 [0.117, 0.120] & 0.218 [0.215, 0.221] \\
2.00 & 0.115 [0.114, 0.116] & 0.243 [0.239, 0.246] \\
2.25 & 0.112 [0.111, 0.114] & 0.265 [0.262, 0.268] \\
\bottomrule
\end{tabular}
\end{table}

Across guidance 1.75--2.25, the component added by guidance along the current velocity increases motion in that direction by 2.6\% to 3.0\% per unit guidance.
This magnitude change is much smaller than the observed decrease in the lowest-FID gain over the same range.

\paragraph{Population time lag.}
We next measure how the lag-based gain changes with guidance.

\begin{table}[ht]
\centering
\small
\setlength{\tabcolsep}{4pt}
\caption{
Direct scalar population time lag calibration for guided SiT-XL/2 under the public hybrid SDE sampler at NFE 50.
Each candidate uses 1,280 paired paths and all 50 post-update states, including the final Mean step.
Lag is RMS population probe-time mismatch multiplied by $10^3$; lower is better.
Gain brackets are paired 95\% intervals from 2,000 whole-trajectory draws with the trained probe and genuine baseline fixed.
The same paths select and evaluate each scalar; the genuine baseline remains the class-conditional training path.
Lowest-FID gains come from the existing shared 50,000-image grid at guidance 1.75, 2, and 2.25; other entries are not reported (---).
}
\label{tab:cfg-lag-calibration}
\begin{tabular}{@{}rcrrr@{}}
\toprule
Guidance & Lag-based gain [95\%] & Unscaled lag RMS & Scalar lag RMS & Lowest-FID gain \\
\midrule
1.50 & $1.030\;[1.029,1.031]$ & 9.8 & 2.2 & --- \\
1.75 & $1.027\;[1.025,1.028]$ & 8.6 & 1.5 & 1.000 \\
2.00 & $1.024\;[1.021,1.028]$ & 7.6 & 2.1 & 0.975 \\
2.25 & $1.021\;[1.018,1.025]$ & 7.2 & 3.5 & 0.950 \\
2.50 & $1.017\;[1.014,1.018]$ & 7.2 & 4.9 & --- \\
4.00 & $1.003\;[1.000,1.010]$ & 14.8 & 14.7 & --- \\
\bottomrule
\end{tabular}
\end{table}

Every fitted scalar reduces population time lag at its guidance strength.
From guidance 1.75 to 2.25, the lag-based gain falls from 1.027 to 1.021.
The lowest-FID gain falls much faster, from 1.000 to 0.950, and crosses below one while the lag-based gain remains above one.

\paragraph{Endpoint covariance allocation.}
Linear interpolation across the measured gain grids maps the lag-based gains to power slopes 1.020, 1.025, and 1.031 at guidance 1.75, 2, and 2.25.
The lowest-FID settings instead have power slopes 0.993, 0.978, and 0.963.
The lag-based gains give power slopes near one, while FID favors progressively less sharpening as guidance rises.

\Needspace{18\baselineskip}
\paragraph{Precision--recall trade-off.}
We finally measure how reducing gain from one to 0.95 changes precision and recall.

\begin{table}[ht]
    \centering
    \small
    \setlength{\tabcolsep}{3.5pt}
    \caption{
        Precision and recall when reducing constant gain from 1 to 0.95 under classifier-free guidance with the SiT-XL/2 public hybrid SDE sampler at NFE 50.
        Each gain setting contains 50,000 ImageNet-256 images paired by label, initial latent, Brownian stream, and image row; both metrics lie on the zero-to-one scale and higher is better.
        Each signed contrast is the gain-0.95 value minus the gain-1 value and is computed from unrounded metric values.
        Brackets give paired normal 95\% jackknife intervals from twenty common 2,500-row delete groups with the fixed 10,000-image real reference.
    }%
    \label{tab:guided-precision-recall}
            \begin{tabular}{@{}rrrr@{}}
\toprule
CFG $\omega$ & \shortstack{Precision\\gain 1} & \shortstack{Precision\\gain 0.95} & $\Delta$precision [95\%] \\
\midrule
1.75 & 0.845 & 0.797 & $-0.047\;[-0.051,-0.044]$ \\
2.00 & 0.871 & 0.837 & $-0.035\;[-0.038,-0.032]$ \\
2.25 & 0.888 & 0.862 & $-0.026\;[-0.028,-0.024]$ \\
\bottomrule
\end{tabular}
\par\vspace{0.7\baselineskip}
\begin{tabular}{@{}rrrr@{}}
\toprule
CFG $\omega$ & \shortstack{Recall\\gain 1} & \shortstack{Recall\\gain 0.95} & $\Delta$recall [95\%] \\
\midrule
1.75 & 0.559 & 0.564 & $+0.006\;[-0.010,+0.021]$ \\
2.00 & 0.523 & 0.531 & $+0.008\;[-0.010,+0.026]$ \\
2.25 & 0.486 & 0.508 & $+0.022\;[+0.006,+0.038]$ \\
\bottomrule
\end{tabular}
\end{table}
\FloatBarrier{}

The point estimates also show that the same gain reduction lowers precision and raises recall (Table~\ref{tab:guided-precision-recall}).

\paragraph{Strong guidance.}
At guidance 4, we tested constant gain 0.976 directly on 50,000 images paired row by row with identity.
FID fell from 13.0 to 12.6; the paired gain-minus-identity contrast was $-0.4$ with a 95\% interval $[-0.5,-0.3]$.

Together, these results show that guidance changes the lowest-FID gain more than either velocity magnitude or population time lag alone predicts.
The helpful gain and its precision--recall trade-off therefore depend on the guidance setting.

\FloatBarrier{}
\Needspace{12\baselineskip}
\section{Scope and practical use}

For the tested unguided ImageNet models, scalar calibration provides a useful starting gain when decoding or image-metric evaluation is costly.
It selects gain from population time lag, and an image-metric search can improve FID further.

The changed-state response transfers from SiT to the ADM U-Net under deterministic sampling; model-time correction transfers under stochastic sampling.
The CIFAR-10 experiments measure lag and gain curves but do not repeat these causal interventions.
The SiT ODE--SDE comparison changes the complete integration rule, so it does not isolate the effect of stochastic noise.

\end{document}